\documentclass{article}
\usepackage{ijcai23}

\usepackage{times}
\usepackage{soul}
\usepackage{url}
\usepackage[hidelinks]{hyperref}
\usepackage[utf8]{inputenc}
\usepackage[small]{caption}
\usepackage{graphicx}
\usepackage{amsmath}
\usepackage{amsthm}
\usepackage{booktabs}
\usepackage{algorithm}
\usepackage{algorithmic}
\usepackage[switch]{lineno}
\usepackage{graphicx}
\usepackage{amsmath}
\usepackage{amssymb}
\usepackage{booktabs}
\usepackage{color}
\usepackage{epsfig}
\usepackage{graphicx}
\usepackage{amsmath}
\usepackage{verbatim}
\usepackage{amssymb}
\usepackage{multirow}
\usepackage{booktabs}
\usepackage{tabularx}
\usepackage{url}
\usepackage{graphicx}
\usepackage{mathrsfs}
\usepackage{amsmath}
\usepackage{multirow}
\usepackage{rotfloat}
\usepackage{algorithmic}
\usepackage{algorithm}
\usepackage{graphicx}
\usepackage[table,xcdraw]{xcolor}
\usepackage[normalem]{ulem}
\useunder{\uline}{\ul}{}
\usepackage{cuted}

\usepackage{bbding}
\usepackage{pifont}
\usepackage{wasysym}
\usepackage[table,xcdraw]{xcolor}

\title{Bi-level Dynamic Learning  for Jointly Multi-modality Image Fusion and Beyond}

\author{
Zhu Liu$^1$
\and
Jinyuan Liu$^2$\and
Guanyao Wu$^1$\and
Long Ma$^1$\and
Xin Fan$^3$\And
Risheng Liu$^{3}$\thanks{Corresponding author}
\affiliations
$^1$School of Software Technology, Dalian University of Technology, China\\
$^2$School of Mechanical Engineering, Dalian University of Technology, China\\
$^3$International School of Information Science Engineering, Dalian University of Technology, China
\emails
liuzhu@mail.dlut.edu.cn,
atlantis918@hotmail.com,
rsliu@dlut.edu.cn
}

\begin{document}

\maketitle
\thispagestyle{empty}

%

\begin{abstract}
Recently, multi-modality scene perception tasks, e.g.,  image fusion and scene understanding, have attracted widespread attention for intelligent vision systems. However, early efforts always consider boosting a single task unilaterally and neglecting others, seldom investigating their underlying connections for joint promotion. To overcome these limitations, we establish the hierarchical dual tasks-driven deep model to bridge these tasks. Concretely, we firstly construct an image fusion module to fuse complementary characteristics and cascade dual task-related modules, including a discriminator for visual effects and a semantic network for feature measurement. 
We provide a  bi-level perspective to formulate image fusion and follow-up downstream tasks. To incorporate distinct task-related responses for image fusion, we consider image fusion as a primary goal and dual modules as learnable constraints. Furthermore, we develop an efficient first-order approximation to compute corresponding gradients and present dynamic weighted aggregation to balance the gradients for fusion learning. Extensive experiments demonstrate the superiority of our method, which not only produces visually pleasant fused results but also realizes significant promotion for detection and segmentation than the state-of-the-art approaches.
\end{abstract}

\section{Introduction}
\label{sec:intro}
In real-world scenarios, multi-sensor vision systems play a fundamental role in intelligent applications, \emph{e.g.,}  autonomous driving and robotics. With the rapid deployment of sensor hardware, how utilizing multi-modality images to provide comprehensive scene parsing become an urgent issue~\cite{liu2020bilevel,jiang2022towards,liu2021searching}. Among them, infrared and visible sensors are two widely-used devices, aiming to capture the complete scene information. In detail, visible images can effectively describe the texture details but are sensitive to illumination changes. Infrared images can effectively highlight thermal targets  but lack texture details. Thus, combining the diverse modalities into one image to preserve complementary characteristics is a significant way for both observation~\cite{U2Fusion2020,abs-2303-06840} and  understanding~\cite{li2022rgb,ZhaoZXLP22,ma2022toward}. 
\begin{figure*}
	\centering
	\setlength{\tabcolsep}{1pt} 
	
	\includegraphics[width=0.97\textwidth,]{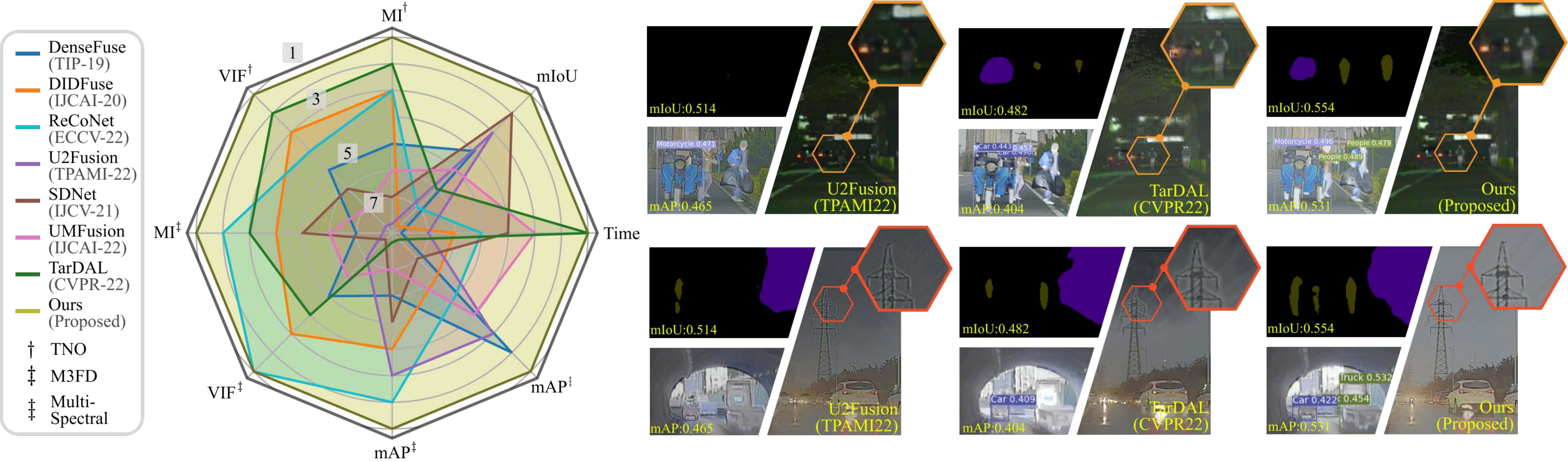}

	\caption{Comprehensive comparisons with advanced multi-modality image fusion methods on visual quality, object detection, and  segmentation. The left figure plots the rankings on diverse metrics with four datasets. The right figures depict the  visual comparisons.}
	
	\label{fig:firstfig}
\end{figure*}

Benefiting from the effective feature extraction of deep learning, various learning-based image fusion approaches have been developed, which can be divided into two mainstream categories, \emph{i.e.,}  auto-encoders~\cite{MFEIF2021,li2018densefuse} and 
end-to-end learning approaches~\cite{U2Fusion2020,zhang2021sdnet,liu2023holoco}. As for auto-encoders, $\ell_{1}$ norm~\cite{li2018densefuse}, weighted average~\cite{MFEIF2021} and maxing choose~\cite{zhao2020didfuse} are widely used for fuse features from pretrained auto-encoders.  Introducing diverse statistic measurements (\emph{e.g.,}  modal gradients~\cite{zhang2021sdnet} and image quality assessment~\cite{U2Fusion2020}), various end-to-end learning methods are proposed by designing effective architectures. However, we argue that most fusion methods focus on visual quality, ignoring the role to facilitate the downstream perception tasks. Constrained by the visual statistics,  typical features, both benefit for visual quality and perception, are easy to be neglected, causing insufficient learning of both tasks.



Lately, few works~\cite{TarDAL,SeaFusion,sun2022detfusion} attempt to jointly realize the pixel-wise image fusion and semantic perception tasks. These works mostly cascade related networks directly and utilize end-to-end training with multi-task loss functions to realize task-driven image fusion. Unfortunately, there are two shortcomings that limit their performance. (i)~\emph{Lacking the investigation of underlying connection}: Joint learning may place an obstacle to preserving distinct features of tasks and cannot formulate intrinsic mutual-promoted relationships.
(ii)~\emph{Inflexible trade-off of multi-task learning:} Existing methods
mostly utilize manual hyper-parameters to balance the diverse loss functions, which cannot guarantee optimal performances for both tasks. 
Thus, the primary goal of this paper is to realize comprehensive image fusion, in order to achieve joint promotion both for observation and semantic understanding.

\subsection{Our Contribution}
To mitigate these issues, we develop a generic bi-level dynamic learning paradigm for bridging the relationships jointly between multi-modality image fusion and semantic perception tasks.  Concretely, we first establish a hierarchical deep model, composited by an image fusion module,  visual discriminator, and commonly used perception network.
We introduce dual learnable modules for the measurement of visual quality and semantic perception respectively, to provide distinct task-specific responses for image fusion.
More importantly, a bi-level learning paradigm is proposed to formulate the latent connection of hierarchical modules.  We also derive a dynamic weighting aggregation with efficient approximation to realize the mutually reinforcing of visual results and perception jointly. Figure~\ref{fig:firstfig} demonstrates our proposed strategy achieves better visual-appealing fused images and precise semantic perception performance (detection and segmentation) against state-of-the-arts. Our  contributions can be concluded as follows:

\begin{itemize}

	\item  Considering the visual quality and semantic information richness 
	of image fusion as two correlative goals, we propose a hierarchical deep model to realize the mutually reinforced task-driven image fusion.
	\item For the training strategy, we devise a bi-level formulation to bridge the image fusion with two task-specific constraints, providing an efficient way to formulate their inner mutual-benefit relationship. 
	
	\item  For the solving procedure, we drive a dynamic aggregated solution, yielding the efficient gradient approximation and adaptive weighting schemes to balance the gradients from diverse modules automatically,
	for learning the optimal parameters for both tasks.
	
	\item Comprehensive evaluations on three multi-modality vision tasks  (\emph{i.e.,}  image fusion, object detection, and semantic segmentation) are conducted to illustrate the superiority against  state-of-the-art methods.  Sufficient analytical results also substantiate the effectiveness.
	
\end{itemize}

\begin{figure*}[htb]
	\centering
	\setlength{\tabcolsep}{1pt} 
	
	\includegraphics[width=0.97\textwidth,]{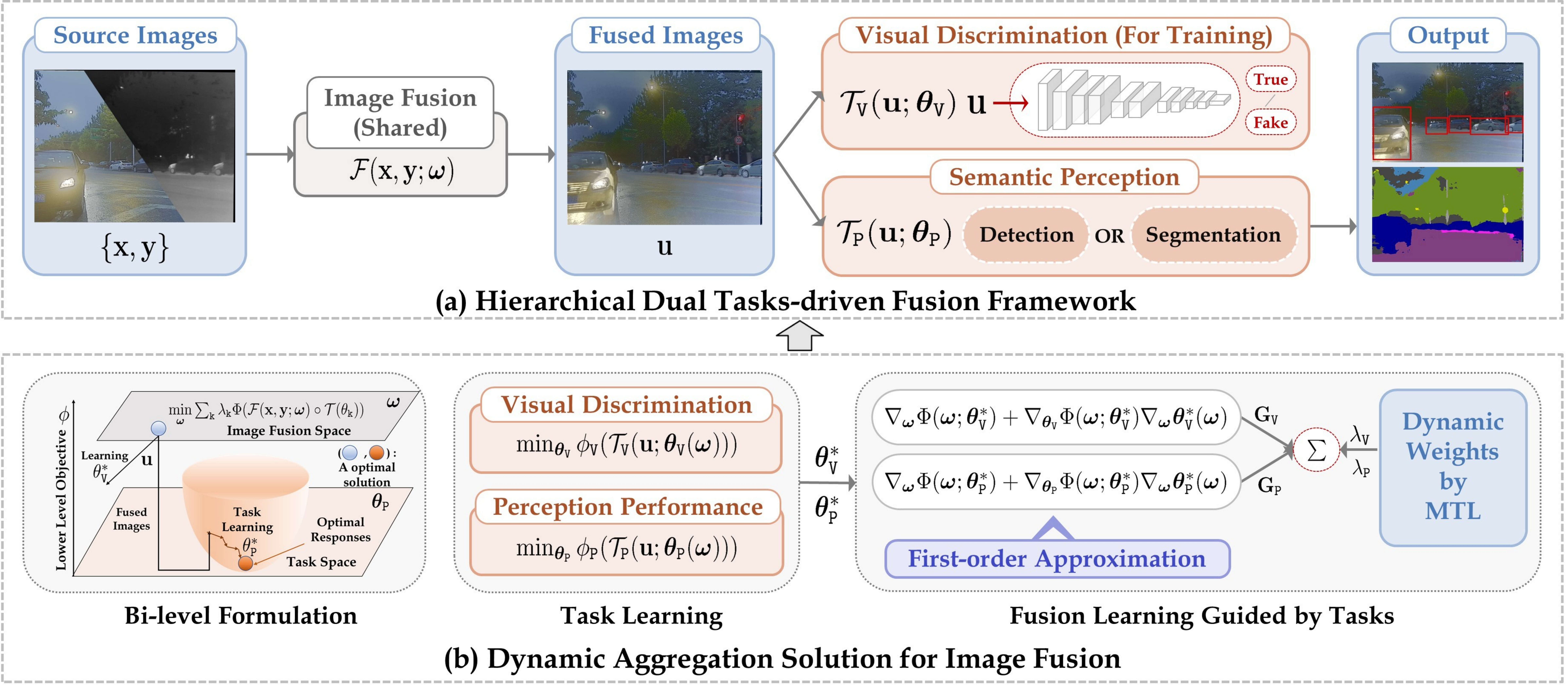}

	\caption{Schematic of the proposed model and strategy. In (a), we plot the hierarchical dual-tasks driven fusion framework on the training and inference phase respectively. The  proposed bi-level formulation with a concrete learning procedure is shown at (b).}
	
	\label{fig:workflow}
\end{figure*}

\section{The Proposed Method}
\subsection{Motivation}
As aforementioned, the most straightforward way is to establish cascaded architectures~\cite{TarDAL,SeaFusion,abs-2211-14461} (denoted as $\mathcal{N}$) for realizing the comprehensive perception. These networks can be directly decomposed as an image fusion module (denoted as $\mathcal{F}$) and a task-oriented module (denoted as $\mathcal{T}$). These mainstream architectures can be formulated as $\mathcal{N} = \mathcal{F} \circ \mathcal{T}$, training by the combination of diverse loss functions. We argue that existing methods rely on handcrafted visual measurements, which are not adaptative and flexible for joint learning. Thus, we propose a hierarchical dual-tasks driven model.
\begin{algorithm}[t] 
	\caption{Dynamic Aggregated Solution.}\label{alg:framework}
	\begin{algorithmic}[1] 
		\REQUIRE Multi-modality inputs $\mathbf{x},\mathbf{y}$, loss function $\phi_\mathtt{V}$, $\phi_\mathtt{P}$, and $\Phi$ and other necessary hyper-parameters.
		
		\STATE Preparing  pairs $\{\mathbf{x},\mathbf{y}\}$ with perception labels $\mathbf{z}^{*}$.
		\STATE \% \emph{Warm start phase.} 
		
		\STATE Pretrain the fusion network $\mathcal{F}$ for initializing ${\omega}$.
		
		\WHILE {not converged}
		
		\STATE ${\theta}_\mathtt{V}\leftarrow{\theta}_\mathtt{V}-\nabla_{{\theta}_\mathtt{V}} \phi_\mathtt{V}(\mathcal{T}_\mathtt{V}(\mathbf{u};{\theta}_\mathtt{V}({\omega})))$.
		\STATE ${\theta}_\mathtt{P}\leftarrow{\theta}_\mathtt{P}-\nabla_{{\theta}_\mathtt{P}} \phi_\mathtt{P}(\mathcal{T}_\mathtt{P}(\mathbf{u};{\theta}_\mathtt{P}({\omega})))$.
		
		\STATE Calculate gradient $\mathbf{G}_\mathtt{V}$ and   $\mathbf{G}_\mathtt{P}$ with  Eq.~\eqref{eq:gradient}.
		\STATE Generate $\lambda_\mathtt{V}$ and $\lambda_\mathtt{P}$ by using RLW.
		\STATE ${\omega}\leftarrow{\omega}-(\lambda_\mathtt{V}\mathbf{G}_\mathtt{V}+\lambda_\mathtt{P}\mathbf{G}_\mathtt{P})$.
		\ENDWHILE

		\RETURN ${\omega}^{*}$, ${\theta}_\mathtt{P}^{*}$.
	\end{algorithmic}
\end{algorithm}
In detail, as for image fusion, we introduce two dense residual blocks~\cite{yan2019attention} to composite the  network $\mathcal{F}$ with parameters ${\omega}$, to maintain the  complementary characteristics from source images for generating fused images $\mathbf{u}$. 
Supposing infrared and visible images as $\mathbf{x}$ and $\mathbf{y}$ with  gray-scale, the fusion learning can be written as $\mathbf{u} = \mathcal{F}(\mathbf{x},\mathbf{y};{\omega})$. In order to measure the adaptive intensity distribution, we introduce a discriminator as $\mathcal{T}_\mathtt{V}$ with parameters ${\theta}_\mathtt{V}$ to measure the texture similarity with source images. Denoted the classified output as $\mathbf{z}_\mathtt{V}$, the discrimination can be formulated as $\mathbf{z}_\mathtt{V} = \mathcal{T}_\mathtt{V} (\mathbf{u};{\theta}_\mathtt{V})$. This formulation can provide adaptive learnable responses, compared with handcrafted loss functions.

Moreover, as for the semantic understanding, two representative networks for object detection~\cite{tian2019fcos} and semantic segmentation~\cite{xie2021segformer}) are selected as the task module $\mathcal{T}_\mathtt{P}$ with parameters ${\theta}_\mathtt{P}$.  Similarly, the task solution can be written as $\mathbf{z}_\mathtt{P} = \mathcal{T}_\mathtt{P}(\mathbf{u};{\theta}_\mathtt{P})$.
Thus, the complete cascaded architecture can be formulated as:

\begin{eqnarray}
\left\{
\begin{aligned}
\mathbf{u} & = \mathcal{F}(\mathbf{x},\mathbf{y};{\omega}),\\
\mathbf{z}_\mathtt{k} & =  \mathcal{T}_\mathtt{k}(\mathbf{u};{\theta}_\mathtt{k}),\\
\end{aligned}
\right.
\end{eqnarray}
where $\mathtt{k} \in \{\mathtt{V},\mathtt{P}\}$ and the workflow is shown in Figure~\ref{fig:workflow} (a).
Specifically, compared with multi-task learning, the fusion network $\mathcal{F}$ actually plays a role for robust feature extraction (can be viewed as ``encoder''). $\mathcal{T}_\mathtt{V}$ and $\mathcal{T}_\mathtt{P}$ are as task-specific ``decoders'' introduced to learn the ability to discriminate fusion quality and measure the informative richness to support the downstream scene perception.

\subsection{Bi-level Formulation}
Recently, various training strategies are proposed to address the high-level task-driven image fusion, including unrolled end-to-end training~\cite{TarDAL}, separate stage-wise training~\cite{wu2022breaking} and adaptive loop-based training~\cite{SeaFusion}. However, we emphasize that these optimization strategies cannot model the coupled mutual-benefit relationship between visual quality and semantic reinterpretation, which is untoward to balance the influences of distinct tasks. Therefore, designing the learning paradigm to realize the ``Best of Both Worlds'' simultaneously is the core goal of this paper.
In this part, we provide a bi-level formulation to depict the overall optimization procedure, in order to illustrate the mutual collaboration and guidance between visual inspection and semantic
perception. The bi-level learning~\cite{liu2021investigating} can be formulated as: 
\begin{align}
&	\min\limits_{{\omega}} \sum_\mathtt{k} \lambda_\mathtt{k} \Phi_\mathtt{k}(\mathcal{F}(\mathbf{x},\mathbf{y};{\omega})\circ \mathcal{T}_\mathtt{k}(\mathbf{\theta}_\mathtt{k})),\label{eq:main}\\
&	\mbox{ s.t. } \left\{
\begin{aligned}
{\theta}_\mathtt{V}^{*} &= \arg\min_{{\theta}_\mathtt{V}}\phi_\mathtt{V}( \mathcal{T}_\mathtt{V}(\mathbf{u};{\theta}_\mathtt{V}({\omega}))), \\
{\theta}_\mathtt{P}^{*} &= \arg\min_{	{\theta}_\mathtt{P}}\phi_\mathtt{P}( \mathcal{T}_\mathtt{P}(\mathbf{u};{\theta}_\mathtt{P}({\omega}))),
\end{aligned}
\right.	
\label{eq:constraint}
\end{align}
where $\mathbf{u} = \mathcal{F}(\mathbf{x},\mathbf{y};{\omega}^{*})$. $\Phi$ and $\phi$ are the objectives on the validation and training datasets respectively. $\lambda_\mathtt{k}$ represents the dynamic multi-task trade-off parameters.
To be more specific, the primary part is to optimize the fusion network $\mathcal{F}$ for extracting rich features, which can be a benefit for the visual quality and semantic perception, expressed by Eq.~\eqref{eq:main}. Furthermore, the discrimination of visual effects and 
responses of semantic understanding are two vital constraints to provide diverse task-specific information, as shown in Eq.~\eqref{eq:constraint}. 
On the other hand, the hierarchical formulation between Eq.~\eqref{eq:main} and Eq.~\eqref{eq:constraint} are nested with mutual promotion. Fused images $\mathbf{u}$ are the fundamental data dependency for following task learning. Based on the responses from vision tasks, task-driven feedback can assist in the optimization of fusion from downstream vision tasks. 
%
\begin{figure*}[htb]
	\centering
	\setlength{\tabcolsep}{1pt}
	\begin{tabular}{cccccccc}
		\includegraphics[width=0.12\textwidth,height=0.085\textheight]{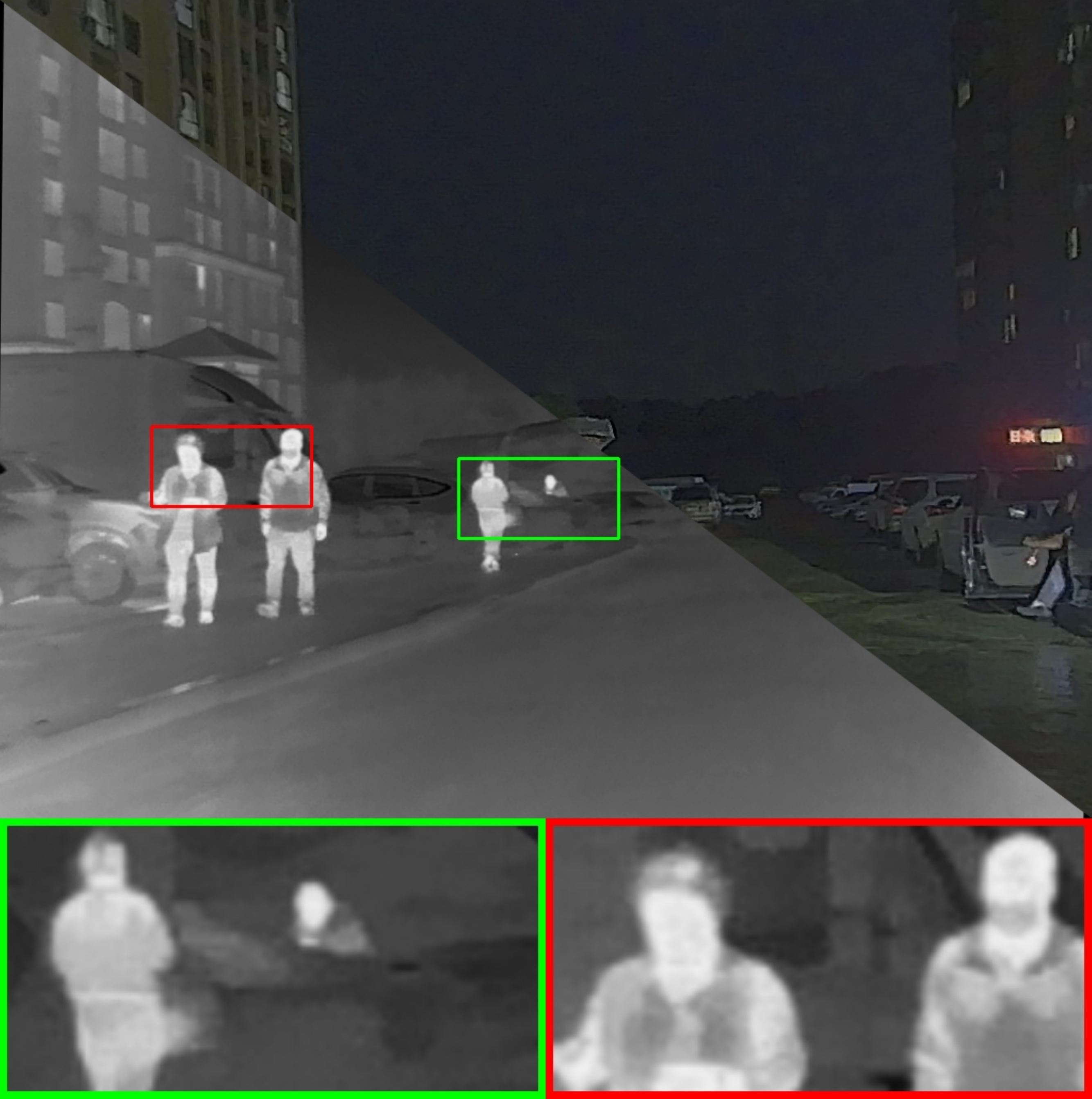}
		&\includegraphics[width=0.12\textwidth,height=0.085\textheight]{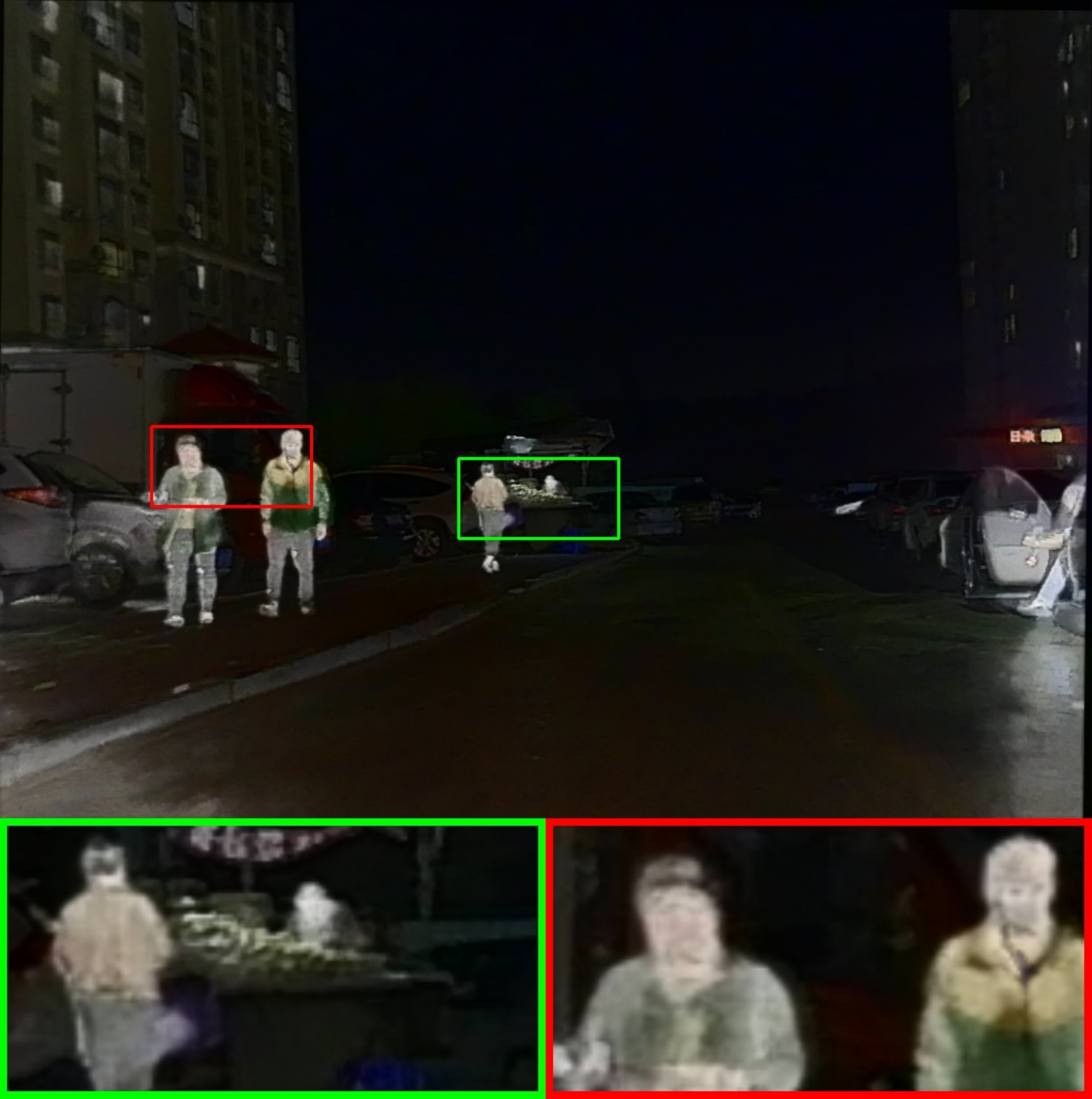}
		&\includegraphics[width=0.12\textwidth,height=0.085\textheight]{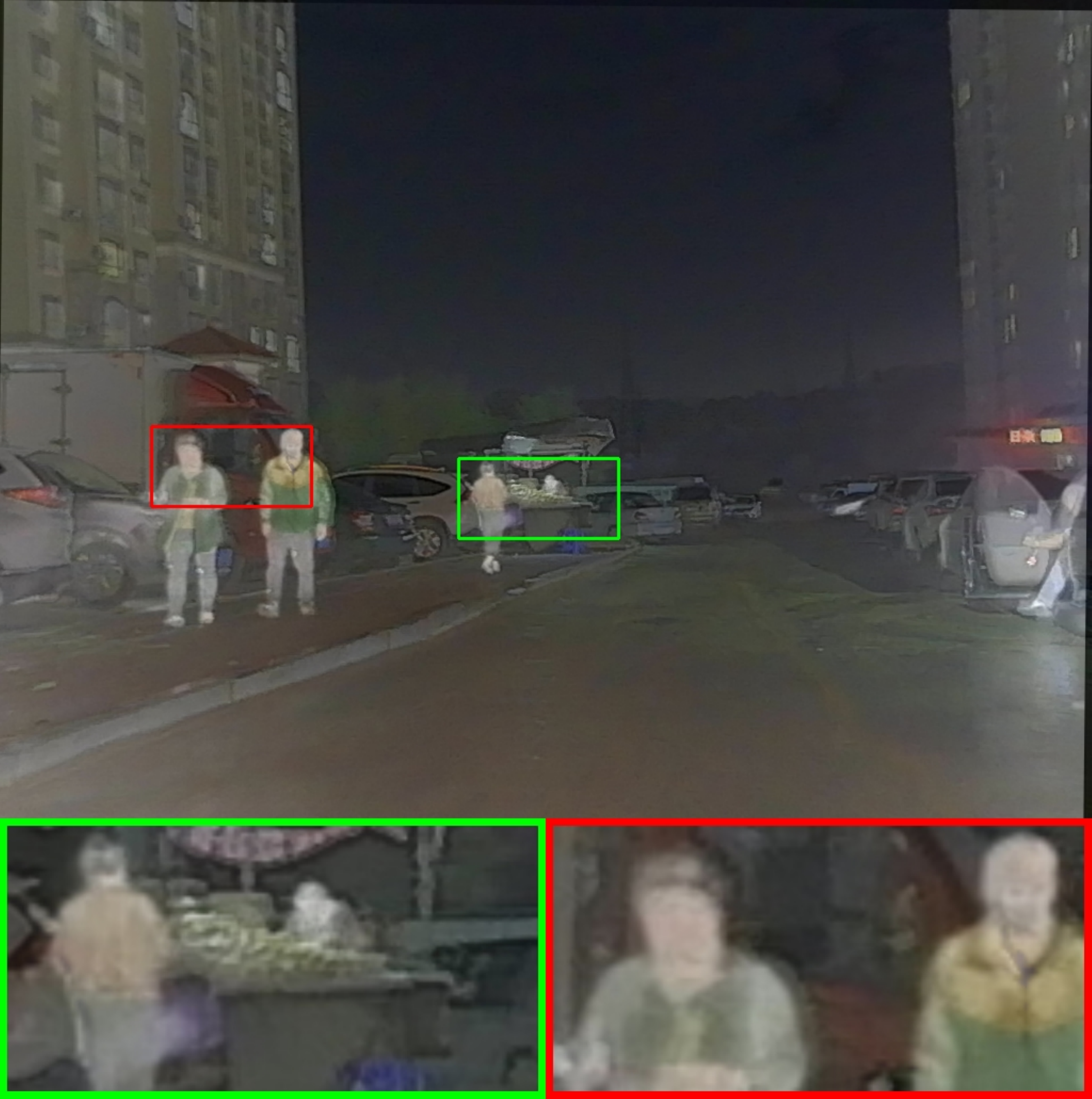}
		&\includegraphics[width=0.12\textwidth,height=0.085\textheight]{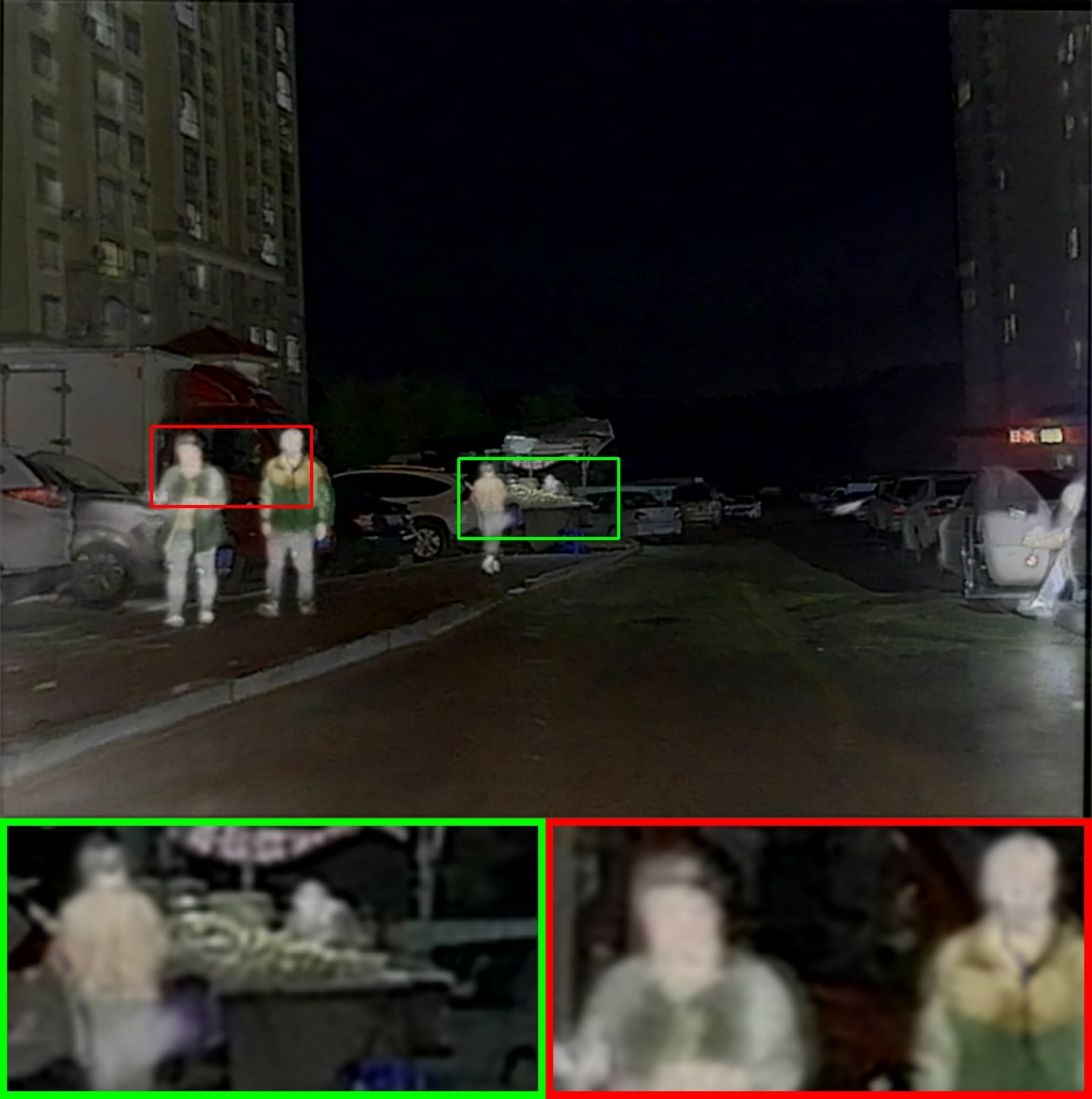}
		&\includegraphics[width=0.12\textwidth,height=0.085\textheight]{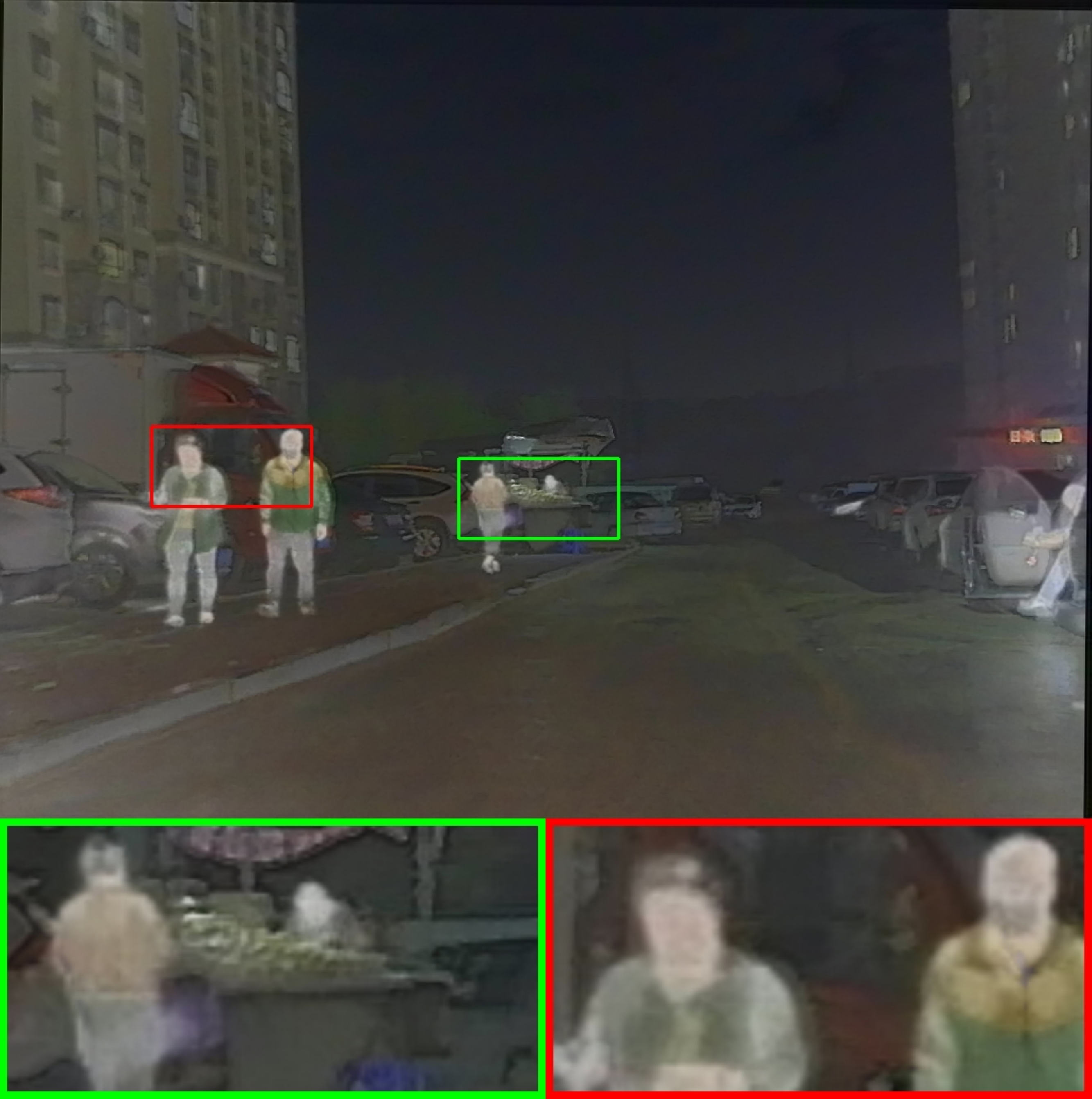}
		&\includegraphics[width=0.12\textwidth,height=0.085\textheight]{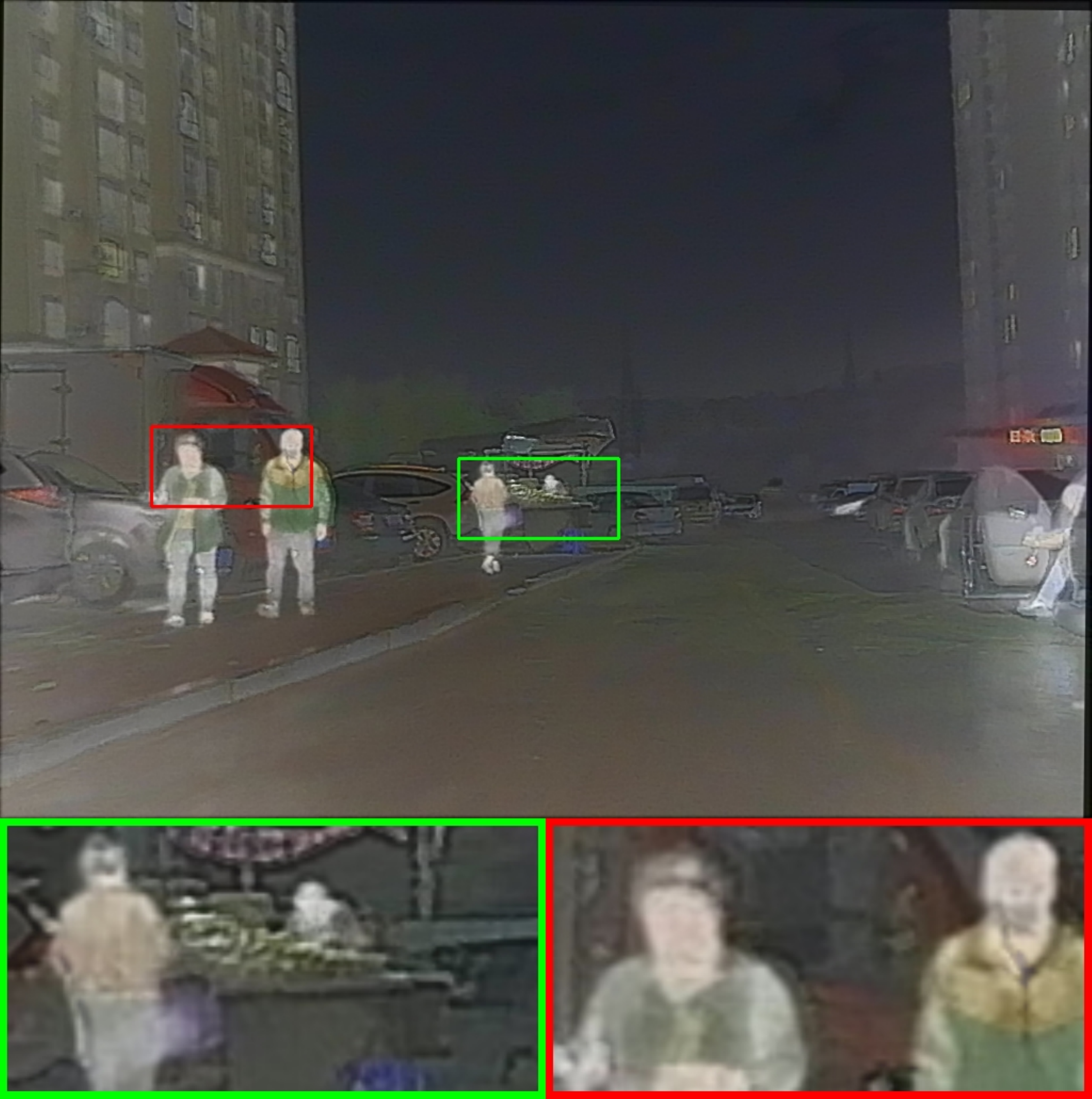}
		&\includegraphics[width=0.12\textwidth,height=0.085\textheight]{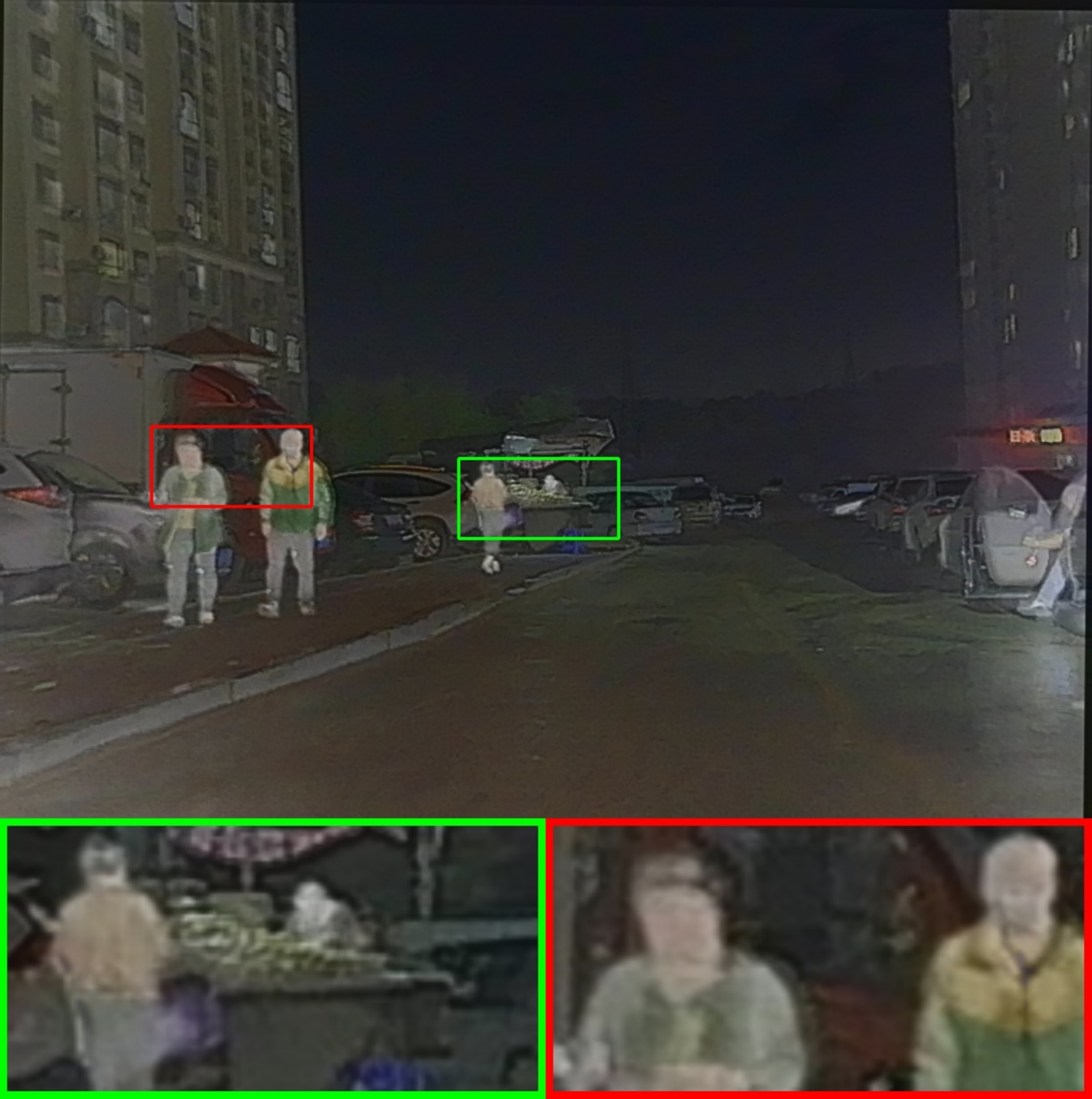}
		&\includegraphics[width=0.12\textwidth,height=0.085\textheight]{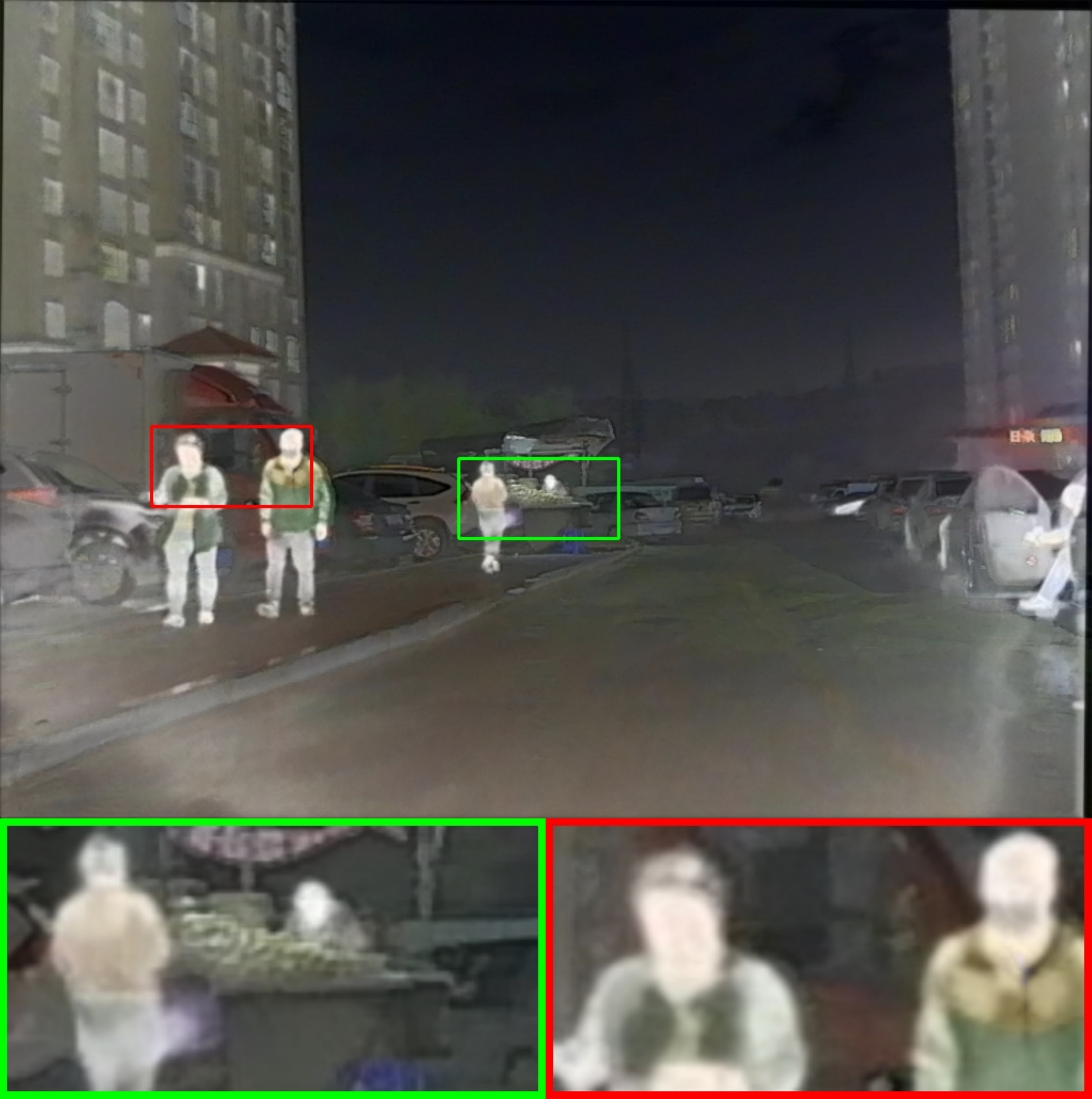}\\
		\includegraphics[width=0.12\textwidth,height=0.085\textheight]{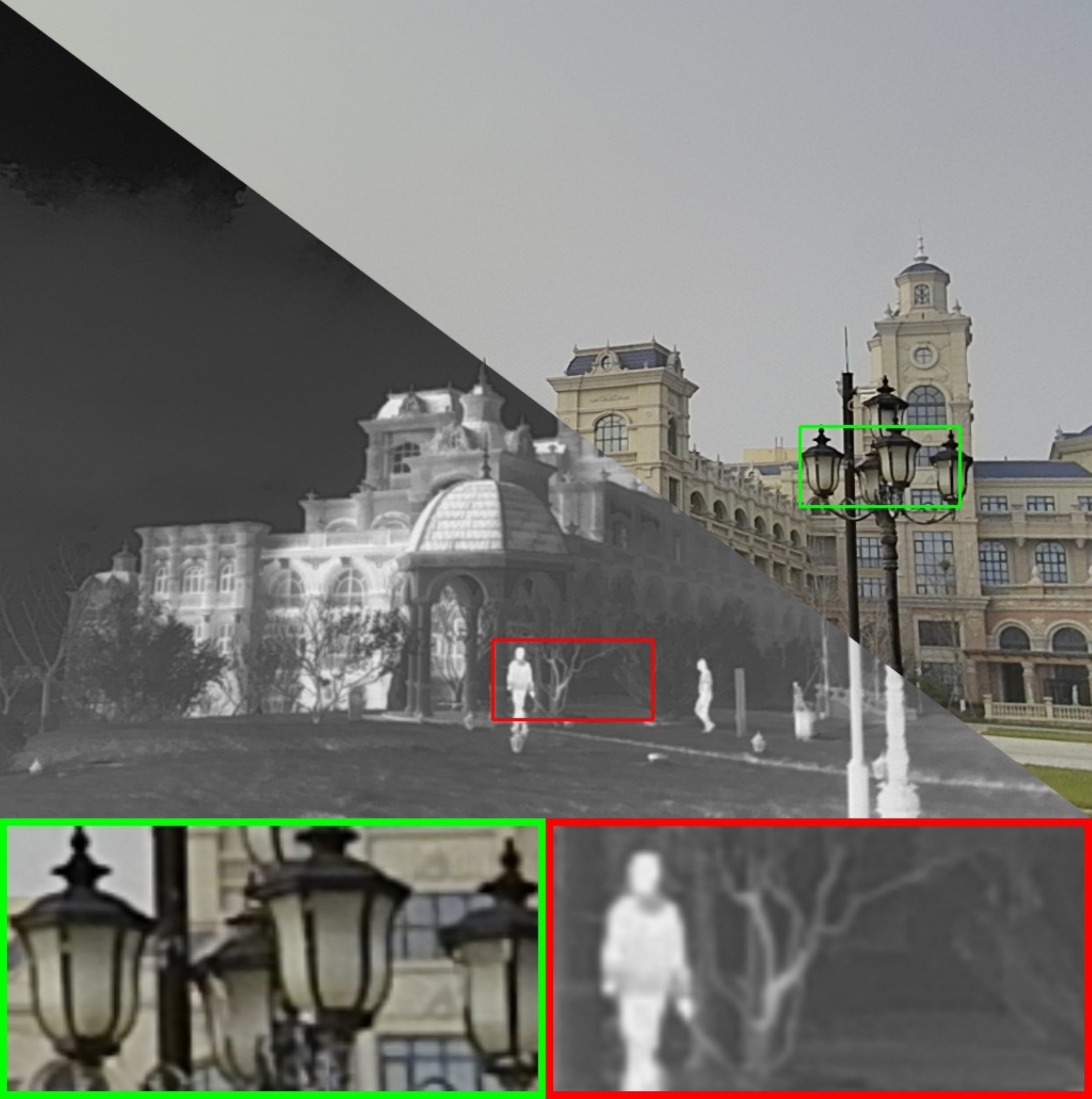}
		&\includegraphics[width=0.12\textwidth,height=0.085\textheight]{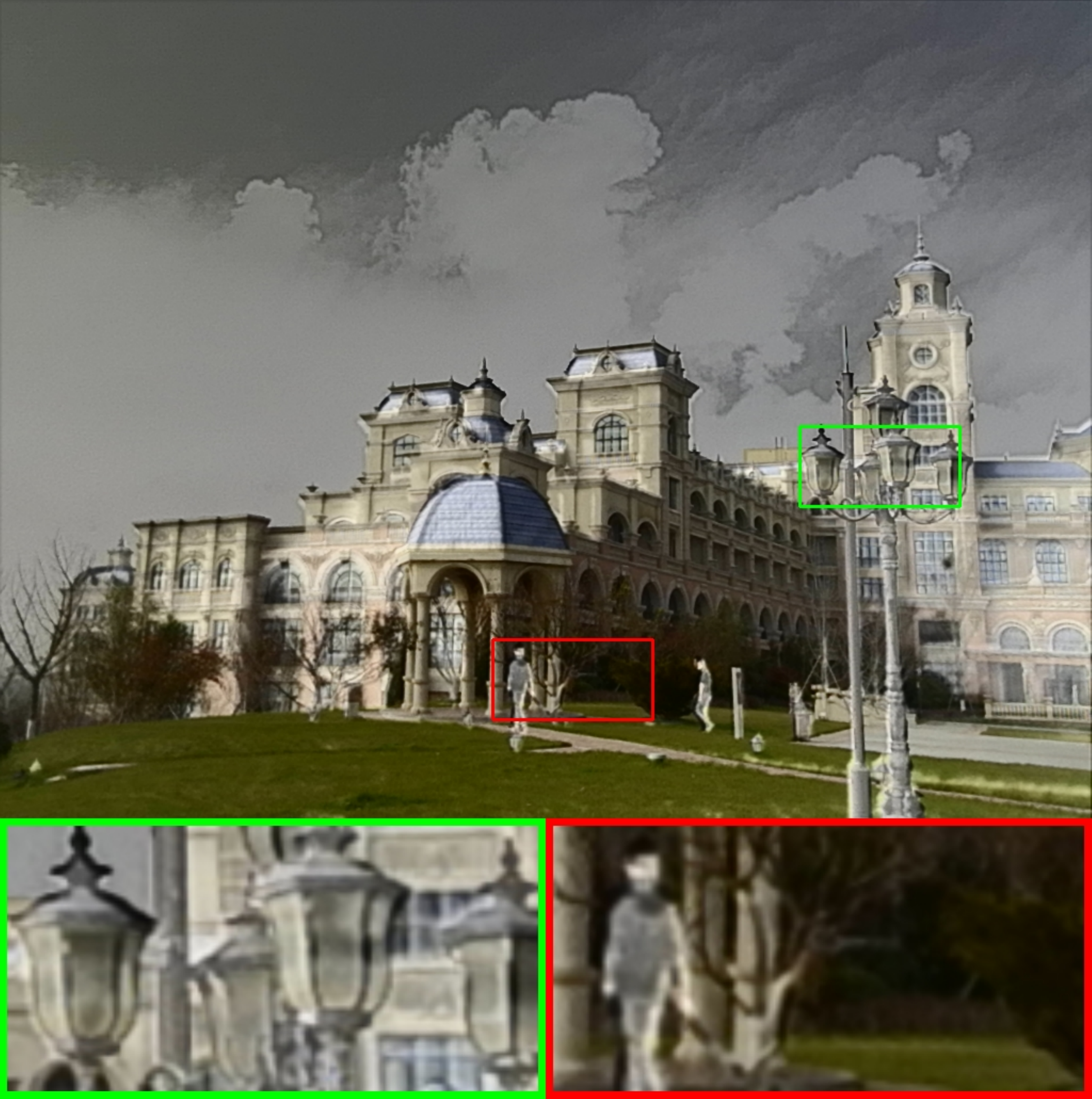}
		&\includegraphics[width=0.12\textwidth,height=0.085\textheight]{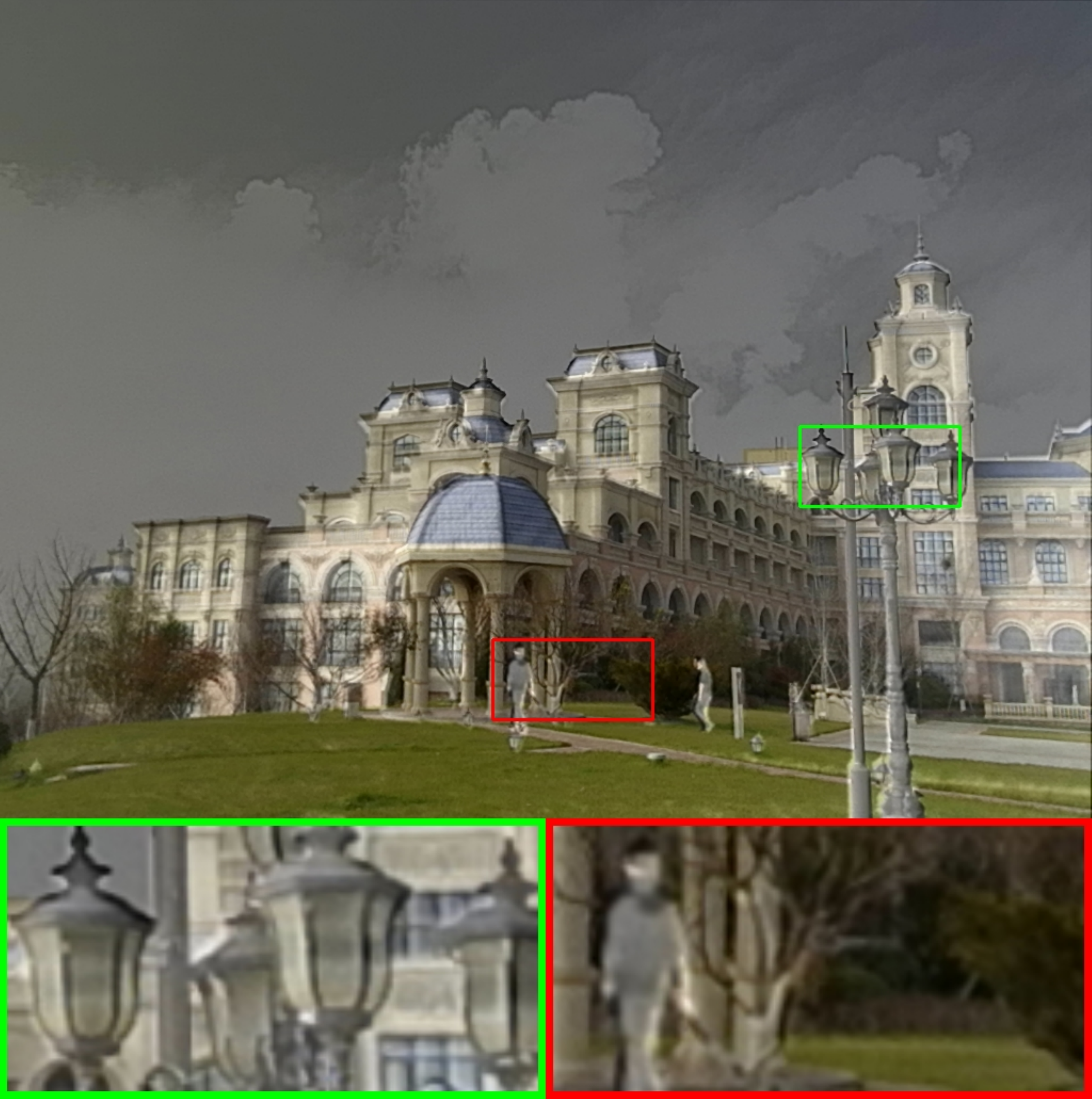}
		&\includegraphics[width=0.12\textwidth,height=0.085\textheight]{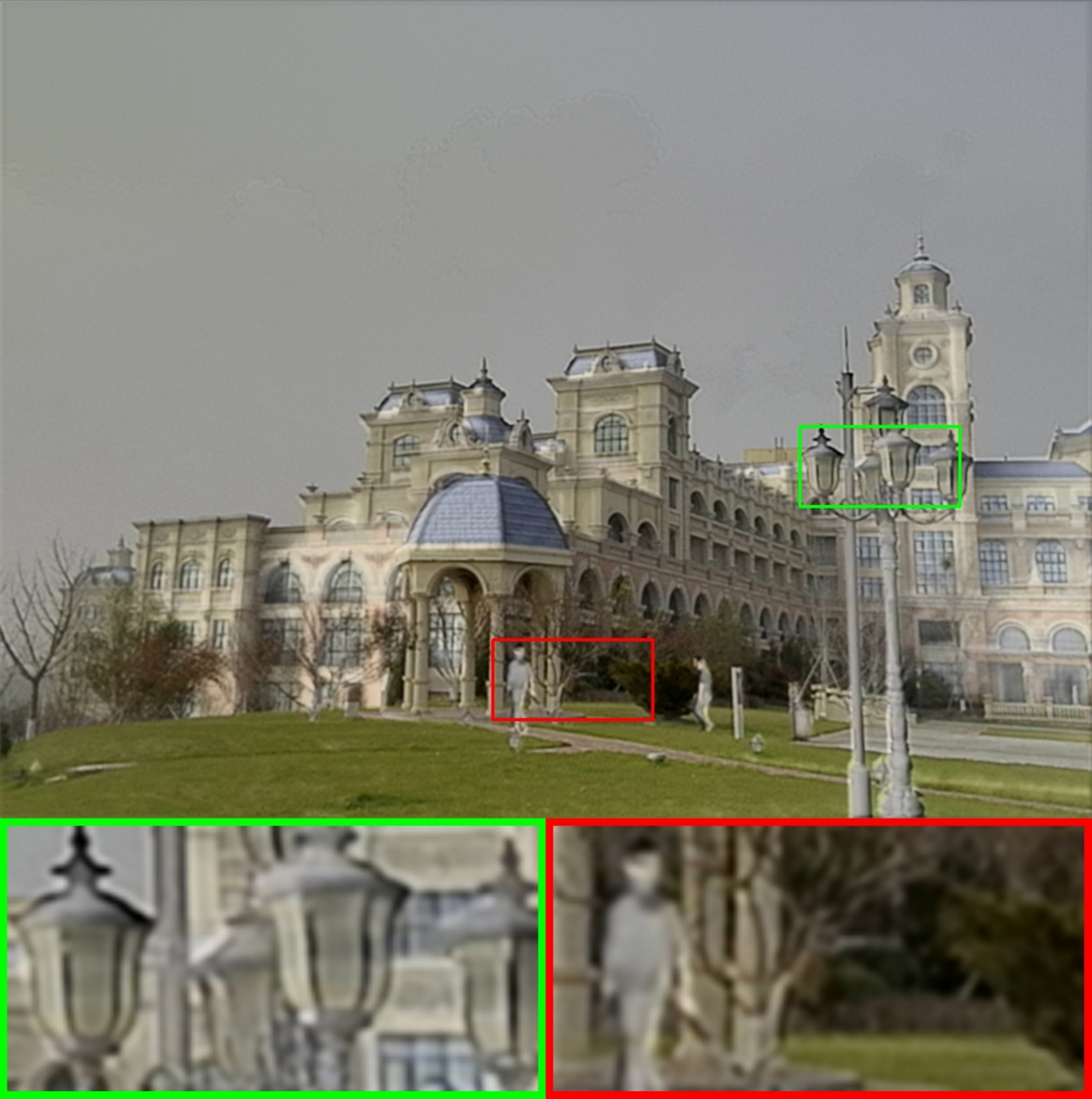}
		&\includegraphics[width=0.12\textwidth,height=0.085\textheight]{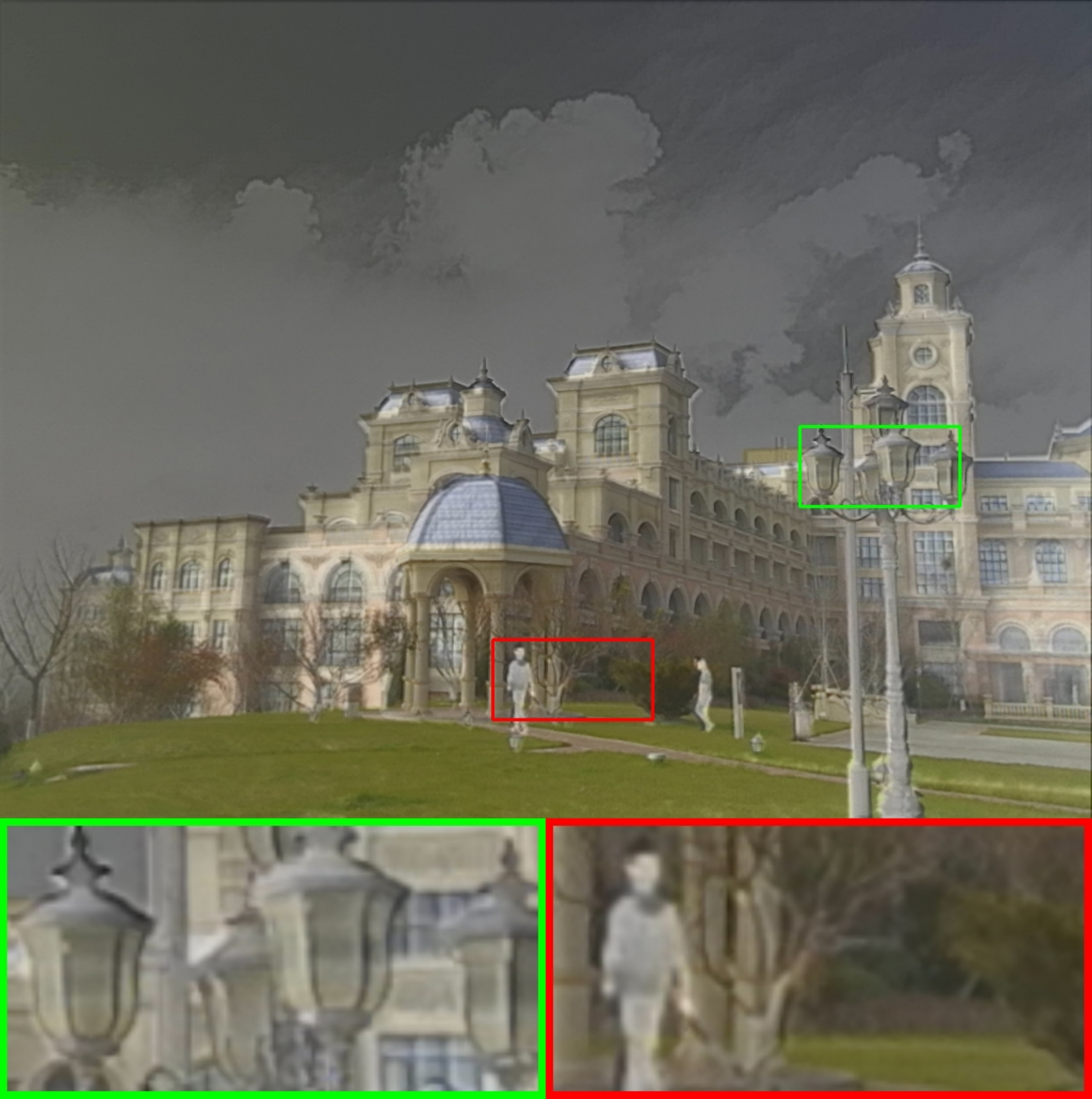}
		&\includegraphics[width=0.12\textwidth,height=0.085\textheight]{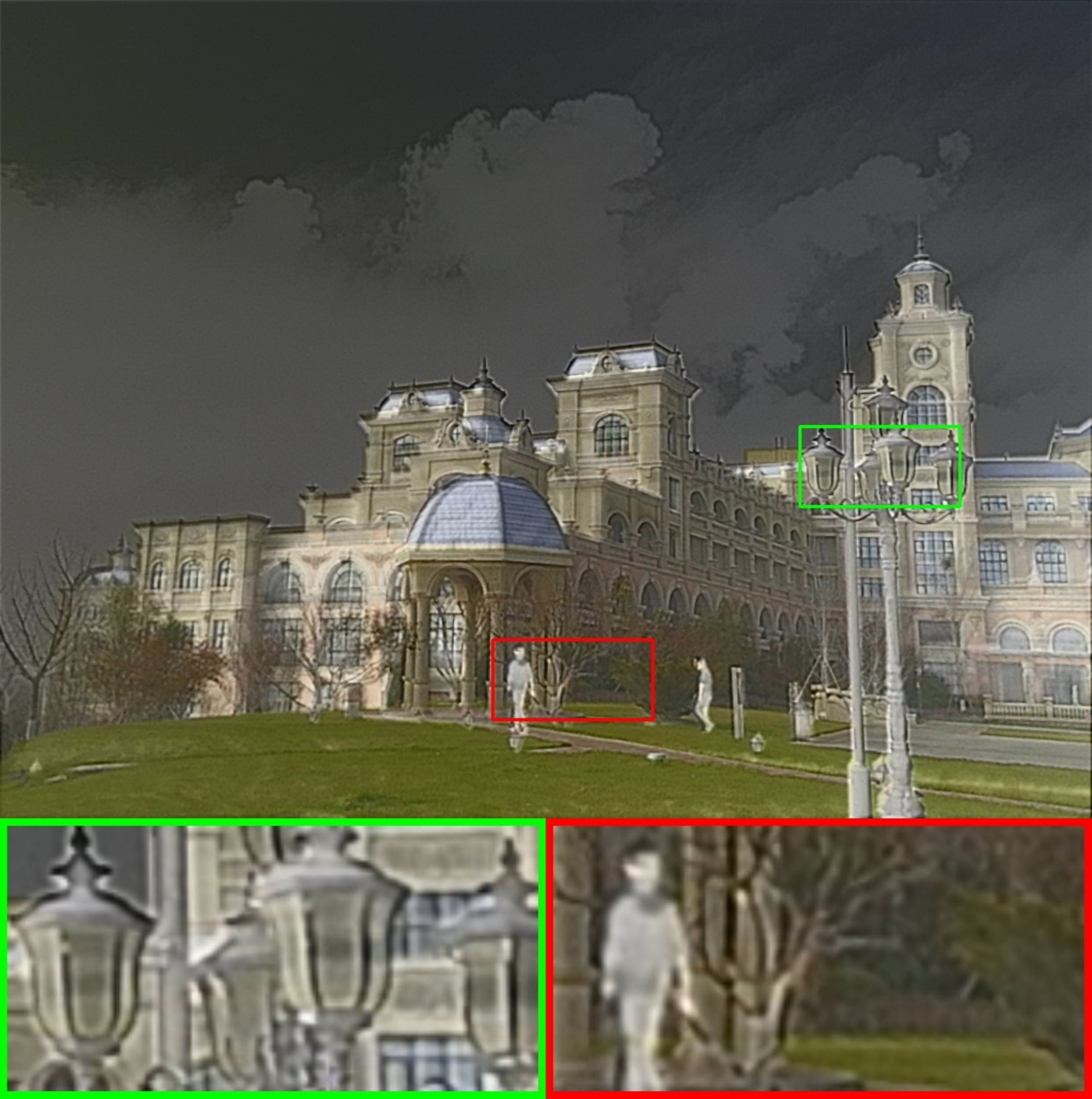}
		&\includegraphics[width=0.12\textwidth,height=0.085\textheight]{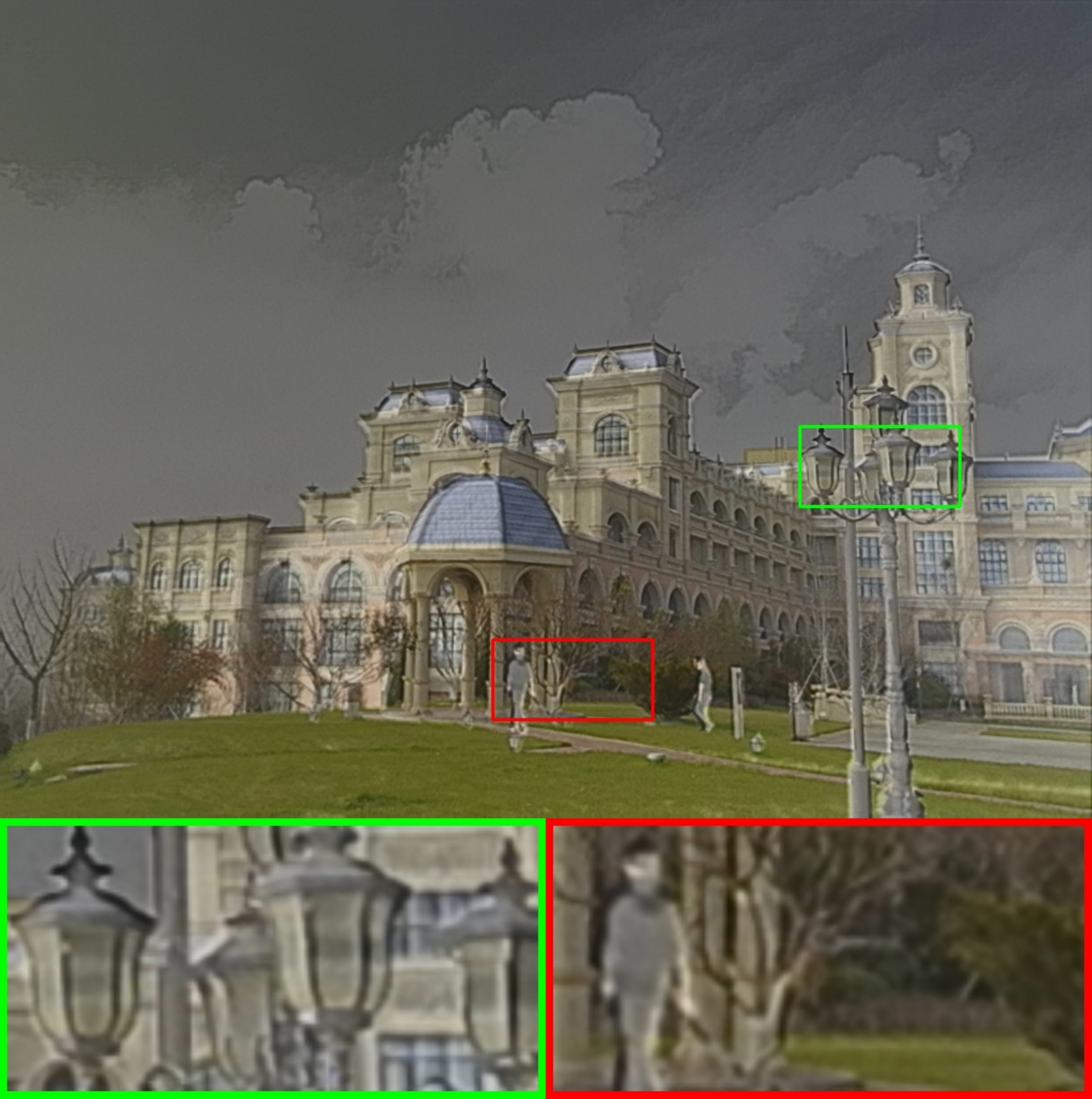}
		&\includegraphics[width=0.12\textwidth,height=0.085\textheight]{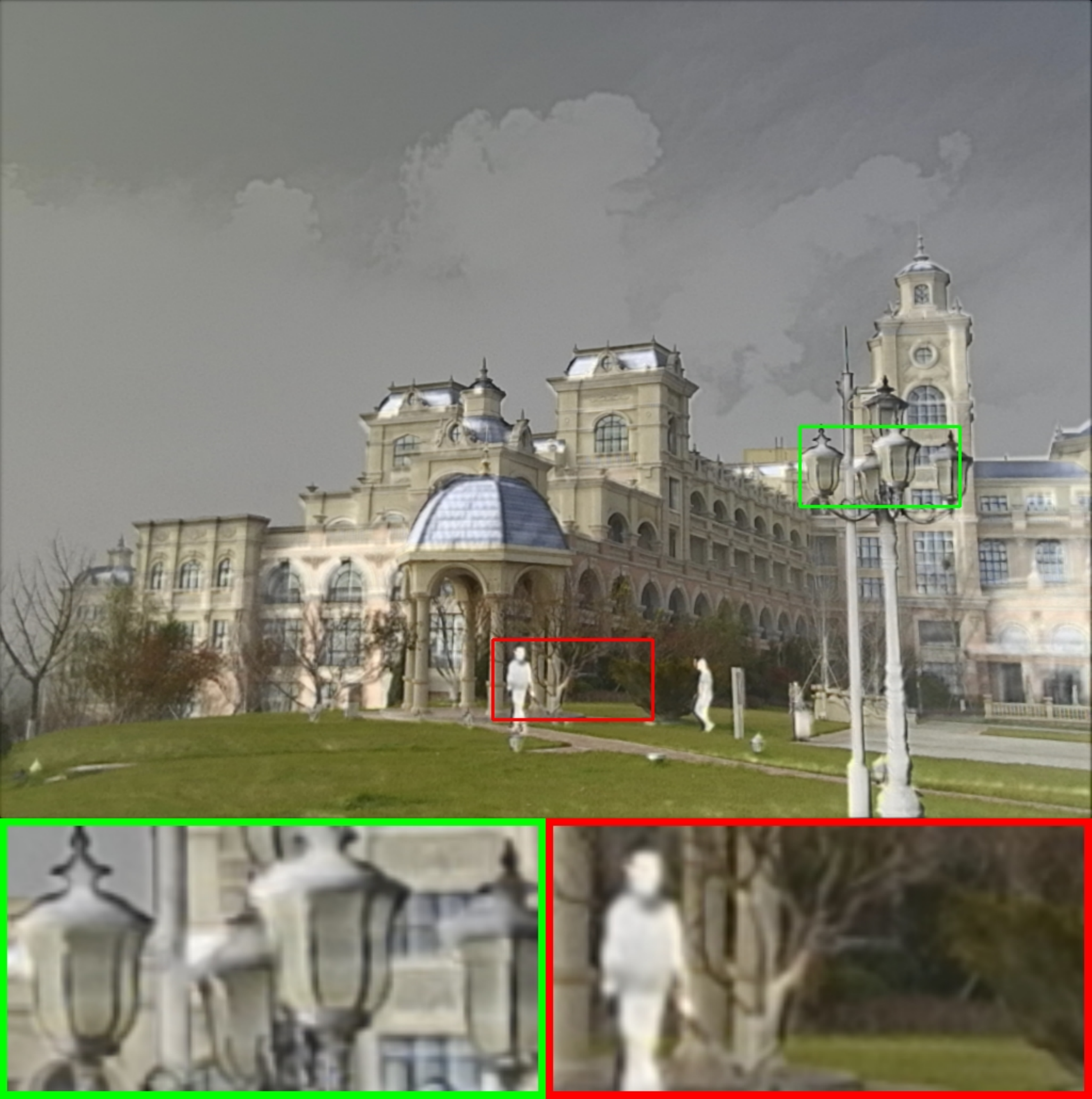}\\
		\footnotesize	Source images&\footnotesize DIDFuse&\footnotesize DenseFuse&\footnotesize ReCoNet&\footnotesize UMFusion&\footnotesize SDNet&\footnotesize U2Fusion&\footnotesize Ours
		\\
	\end{tabular}

	\caption{Visual comparison of different fusion approaches on two challenging scenarios (\emph{i.e.,}  extreme darkness  and small targets).}
	\label{fig:contristive}
\end{figure*}

\begin{table*}[thb]
	\centering
	\footnotesize
	\renewcommand{\arraystretch}{1.1}
	\setlength{\tabcolsep}{1.0mm}{
		\begin{tabular}{c|c|c|c|c|c|c|c|c|c|c|c|c}
			\hline
			Datasets               & Metrics & DDcGAN &  DenseFuse & AUIF & DIDFuse & MFEIF & ReCoNet& UMFusion& SDNet &U2Fusion&  TarDAL &   Ours\\ \hline\hline
			\multirow{3}{*}{TNO} &     \cellcolor{gray!20} {MI$\uparrow$}    & 1.737 & 2.248 & 2.181 & 2.349 & 2.496 &2.349 & 2.071 & 1.952&1.811& \textcolor{blue}{\textbf{2.648}}  &\textcolor{red}{\textbf{2.946}} \\ \cline{2-13} 
			&    \cellcolor{gray!20} {VIF$\uparrow$}     & 0.683 &0.798  &0.819 &0.832  &0.783&0.824&0.695&0.754&0.678&\textcolor{blue}{\textbf{0.860}}&\textcolor{red}{\textbf{0.913}} \\ \cline{2-13} 
			&   \cellcolor{gray!20} {FMI$\uparrow$}      & 0.858 &\textcolor{blue}{\textbf{0.890}}& 0.879 &0.863  &0.891&0.878&0.888&0.883&0.879&0.881&\textcolor{red}{\textbf{0.892}} \\ 
			\hline
			\multirow{3}{*}{RoadScene} &     \cellcolor{gray!20} {MI$\uparrow$}    & 2.569 & 3.044  &3.066 & 3.103  & 3.225 & 3.099& 2.748 & 3.113 & 3.328 & \textcolor{blue}{\textbf{3.391}}  & \textcolor{red}{\textbf{3.526}}   \\ \cline{2-13} 
			&    \cellcolor{gray!20} {VIF$\uparrow$}     &  0.577 & 0.755 &\textcolor{blue}{\textbf{0.842}} &0.793 &0.767&0.750 &0.742 &0.768&0.698& 0.745&\textcolor{red}{\textbf{0.897}}   \\ \cline{2-13} 
			&   \cellcolor{gray!20} {FMI$\uparrow$}      & 0.859  & 0.868 &0.856 &0.853 &\textcolor{blue}{\textbf{0.870}}    & 0.858&0.866&0.863&0.861&0.852&\textcolor{red}{\textbf{0.871}}  \\ 
			\hline
			\multirow{3}{*}{M3FD} &     \cellcolor{gray!20} {MI$\uparrow$}    &2.148& 2.384& 2.399& 2.520& 2.689& \textcolor{blue}{\textbf{2.754}}&2.470& 2.512&2.178&2.558&\textcolor{red}{\textbf{3.033}}  
			
			\\ \cline{2-13} 
			&    \cellcolor{gray!20} {VIF$\uparrow$}     & 0.570&0.600& 0.682& 0.685& 0.645& \textcolor{red}{\textbf{0.699}}  & 0.585& 0.533& 0.563& 0.661&\textcolor{red}{\textbf{0.699}}  
			
			\\ \cline{2-13} 
			&   \cellcolor{gray!20} {FMI$\uparrow$}      & 0.836&\textcolor{red}{\textbf{0.863}}& 0.845&0.831&0.848&0.845&0.855&0.846 &0.850 &0.825& \textcolor{red}{\textbf{0.863}}
			\\ 
			\hline\hline
		\end{tabular}
		
	}

	\caption{ Quantitative  results of visual fusion quality on three representative datasets  with ten competitive methods. }~\label{tab:fusion}
\end{table*}

\subsection{Dynamic Aggregation Solution}
This part details the solution to address the aforementioned bi-level formulation (Eq.~\eqref{eq:main} and Eq.~\eqref{eq:constraint}). In order to accelerate the training convergence, we first introduce a warm-start strategy to pretrain the fusion network. Then, we present 
a dynamic aggregation solution to jointly address fusion and perception. The concrete optimization procedure is shown in Figure~\ref{fig:workflow} (b). It actually can be expressed with hierarchical optimization, \emph{i.e.,}  task learning (Eq.~\eqref{eq:constraint}) and task-guided fusion learning (Eq.~\eqref{eq:main}).
Following with existing practical strategies~\cite{liu2021investigating}, we first optimize the lower-level task constraints with several steps to estimate the optimal parameters ${\theta}_\mathtt{V}^{*}$ and ${\theta}_\mathtt{P}^{*}$, in order to learn the measurement for visual quality and perception based on task-specific losses.

Considering the mutual influence between fusion and lower vision tasks, represented by ${\theta}_\mathtt{k}({\omega}), \mathtt{k}\in\{\mathtt{V},\mathtt{P}\}$, there actually exists
a complicated connection between the hierarchical tasks, which can be used to measure the response of tasks facing with the changes of fused images. As for the optimization of image fusion, dual gradients can be obtained, which can be written as:
\begin{eqnarray}
\left\{
\begin{aligned}
\mathbf{G}_\mathtt{V} & = \nabla_{{\omega}}\Phi_\mathtt{V}({\omega};{\theta}_\mathtt{V}^{*}) + \nabla_{{\theta}_\mathtt{V}}\Phi_\mathtt{V}({\omega};{\theta}_\mathtt{V}^{*})
\nabla_{{\omega}} {\theta}_\mathtt{V}^{*}({\omega}),\\
\mathbf{G}_\mathtt{P} & =  \nabla_{{\omega}}\Phi_\mathtt{P}({\omega};{\theta}_\mathtt{P}^{*}) + \nabla_{{\theta}_\mathtt{P}}\Phi_\mathtt{P}({\omega};{\theta}_\mathtt{P}^{*})
\nabla_{{\omega}} {\theta}_\mathtt{V}^{*}({\omega}),\\
\end{aligned}
\right. \label{eq:gradient}
\end{eqnarray}

In detail, the gradient $\mathbf{G}_\mathtt{V}$ and $\mathbf{G}_\mathtt{P}$ are computed by
$\Phi_\mathtt{k}(\mathcal{F}(\mathbf{x},\mathbf{y};{\omega})\circ \mathcal{T}_\mathtt{k}(\mathbf{\theta}_\mathtt{k}))$. The first term is a direct gradient in term of ${\omega}$ and the second term depicts the latent coupled connection with follow-up perception tasks.

\paragraph{First-order approximation.}
In literature, solving Eq.~\eqref{eq:gradient} is a challenging issue, where the bottleneck is to compute the second-order gradient (the second term). Inspired by
the Gaussian-Newton approximation, which provides a first-order computation to address continuous learning~\cite{zhou2021image} and generative adversarial learning~\cite{liu2022revisiting}, we introduce this strategy to approximate the Hessian in gradients $\mathbf{G}_\mathtt{V}$ and $\mathbf{G}_\mathtt{P}$. Based on the implicit function theory, we can drive that  $
\nabla_{{\omega}}{\theta}({\omega}) = -\nabla_{{\omega},{\theta} }^{2}\Phi({\omega};{\theta})  \nabla_{{\theta},{\theta}}^{2}\Phi({\omega};{\theta})^{-1}$. Gaussian-Newton approximation can covert the  complicated Hessian matrix with the production of first-order vectors, \emph{i.e.,}
\begin{equation}\label{eq:ida}
\nabla_{{\omega}}{\theta}({\omega}) \approx  \frac{\nabla_{{\theta}}\Phi({\omega};{\theta})^{\top} \nabla_{{\theta}}\phi{{\omega}}({\omega};{\theta})}{\nabla_{{\theta}}\phi({\omega};{\theta})^{\top} \nabla_{{\theta}}\phi({\omega};{\theta})}
\nabla_{{\omega}}\phi({\omega};{\theta}).
\end{equation}
\paragraph{Dynamic gradient aggregation.} Another challenging issue is how to adaptively balance the gradients $\mathbf{G}_\mathtt{V}$ 
and $\mathbf{G}_\mathtt{P}$ to jointly optimize the image fusion network $\mathcal{F}$. Recently, Random Loss Weighting (RLW)~\cite{linreasonable} is advanced in Multi-Task Learning (MTL), which can avoid the local minima with higher generalization and comparable performance. We leverage normal distribution $p(\lambda)$ to generate $\lambda_\mathtt{V}$ and $\lambda_\mathtt{P}$, to avoid the focus on one strongly correlated task and ignoring another. The whole solution is summarized in Alg.~\ref{alg:framework}.

\begin{figure*}[htb]
	\centering
	\setlength{\tabcolsep}{1pt}
	\begin{tabular}{ccccccc}
		\includegraphics[width=0.14\textwidth]{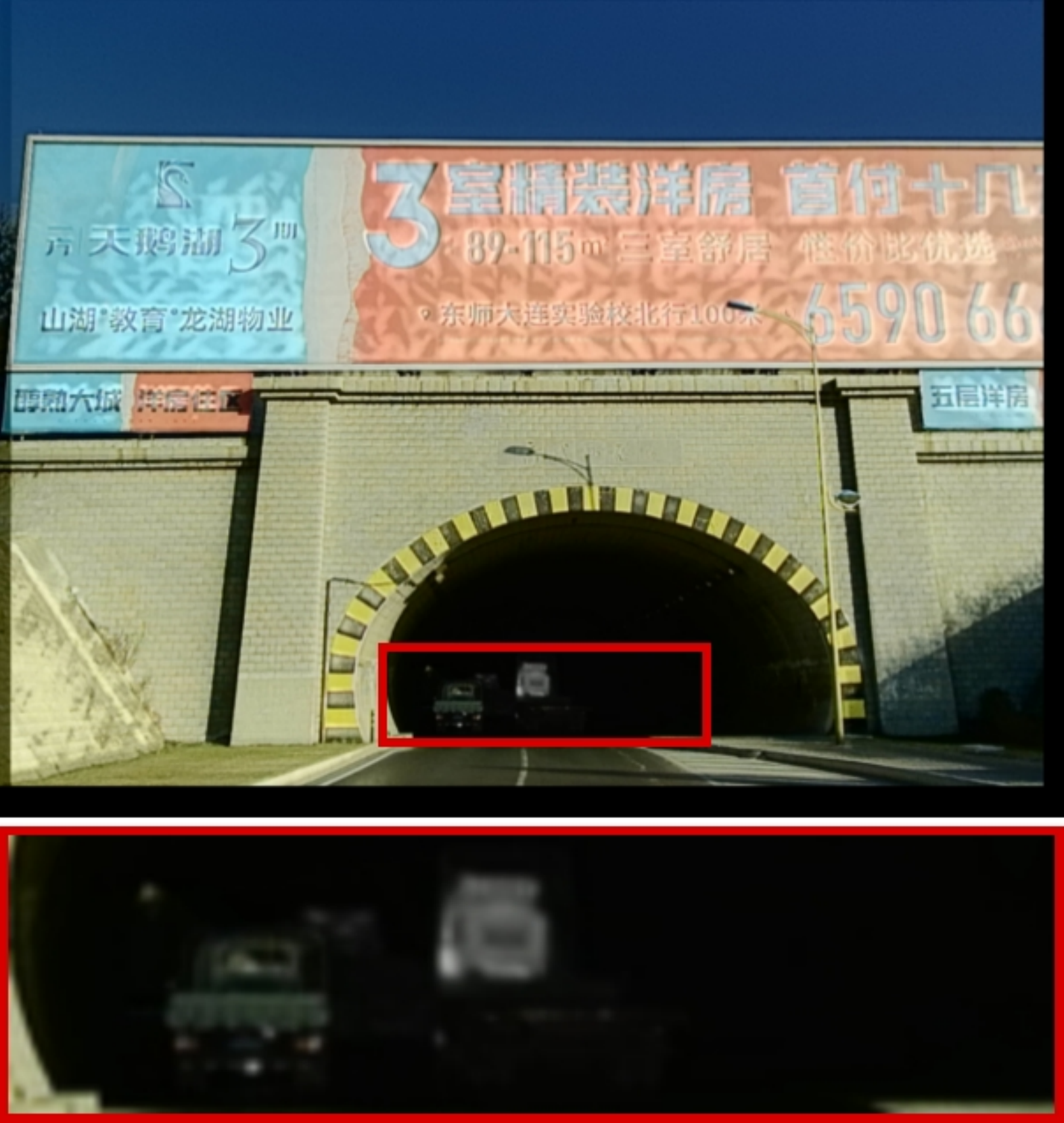}
		&\includegraphics[width=0.14\textwidth]{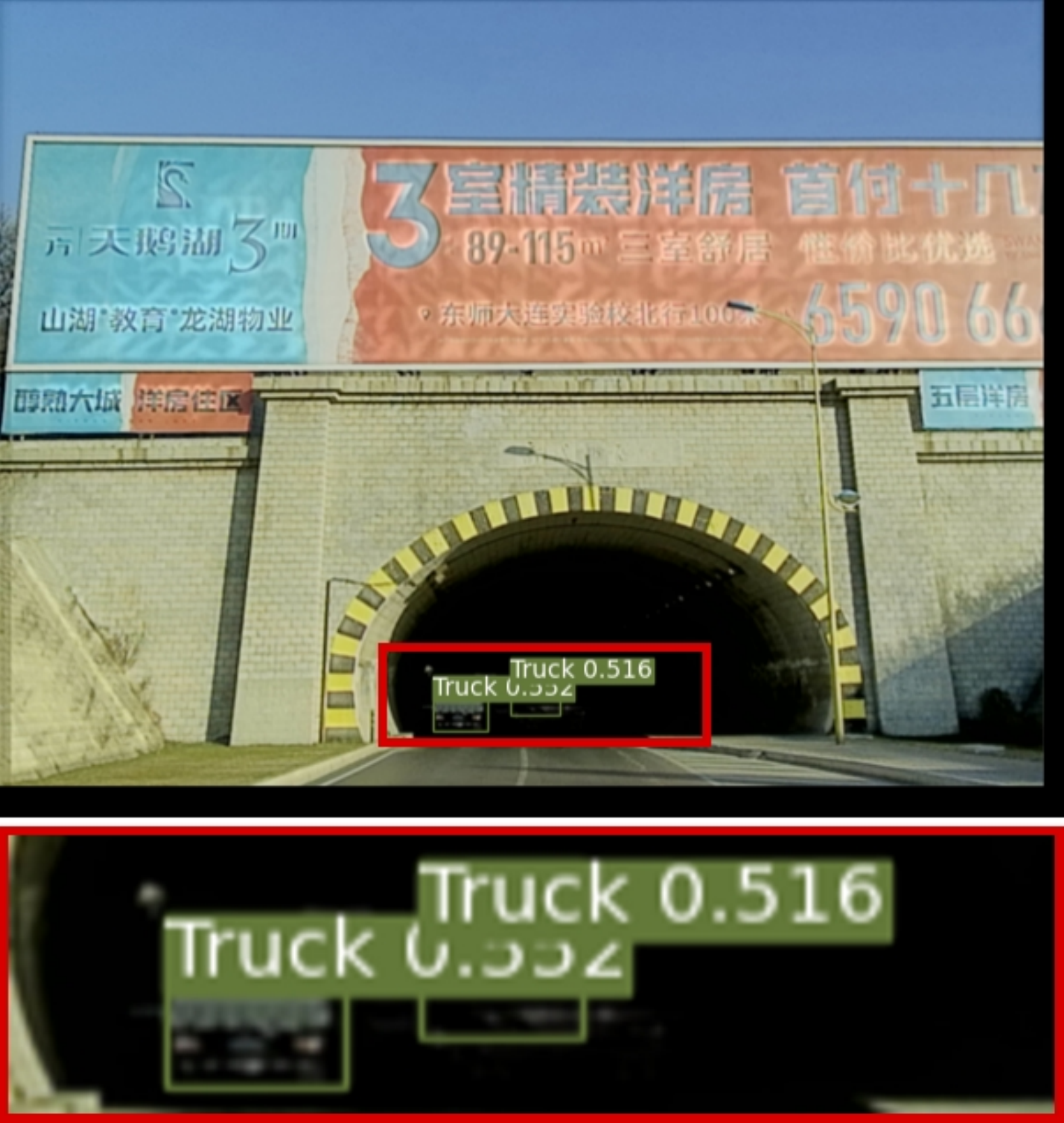}
		&\includegraphics[width=0.14\textwidth]{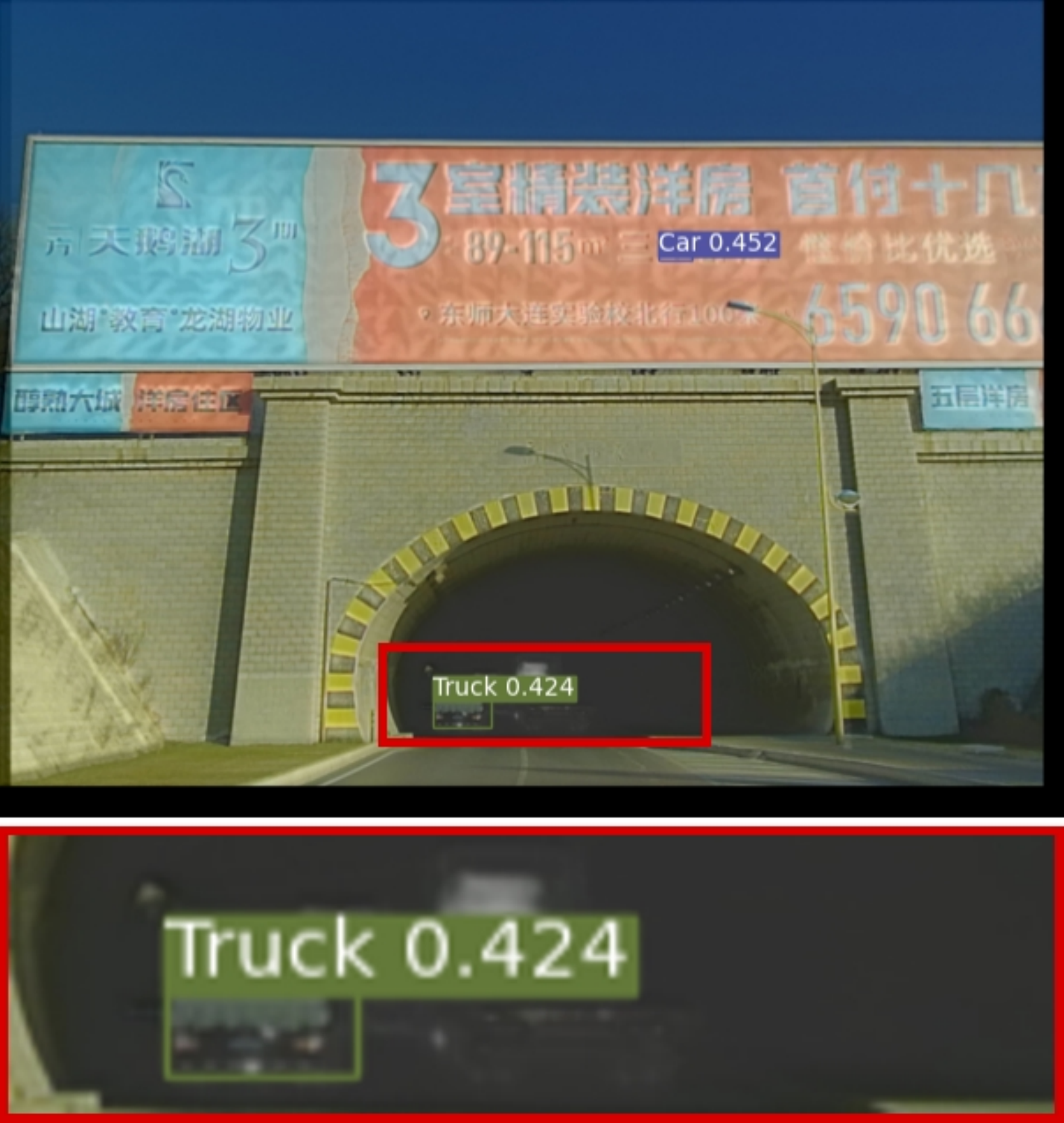}
		&\includegraphics[width=0.14\textwidth]{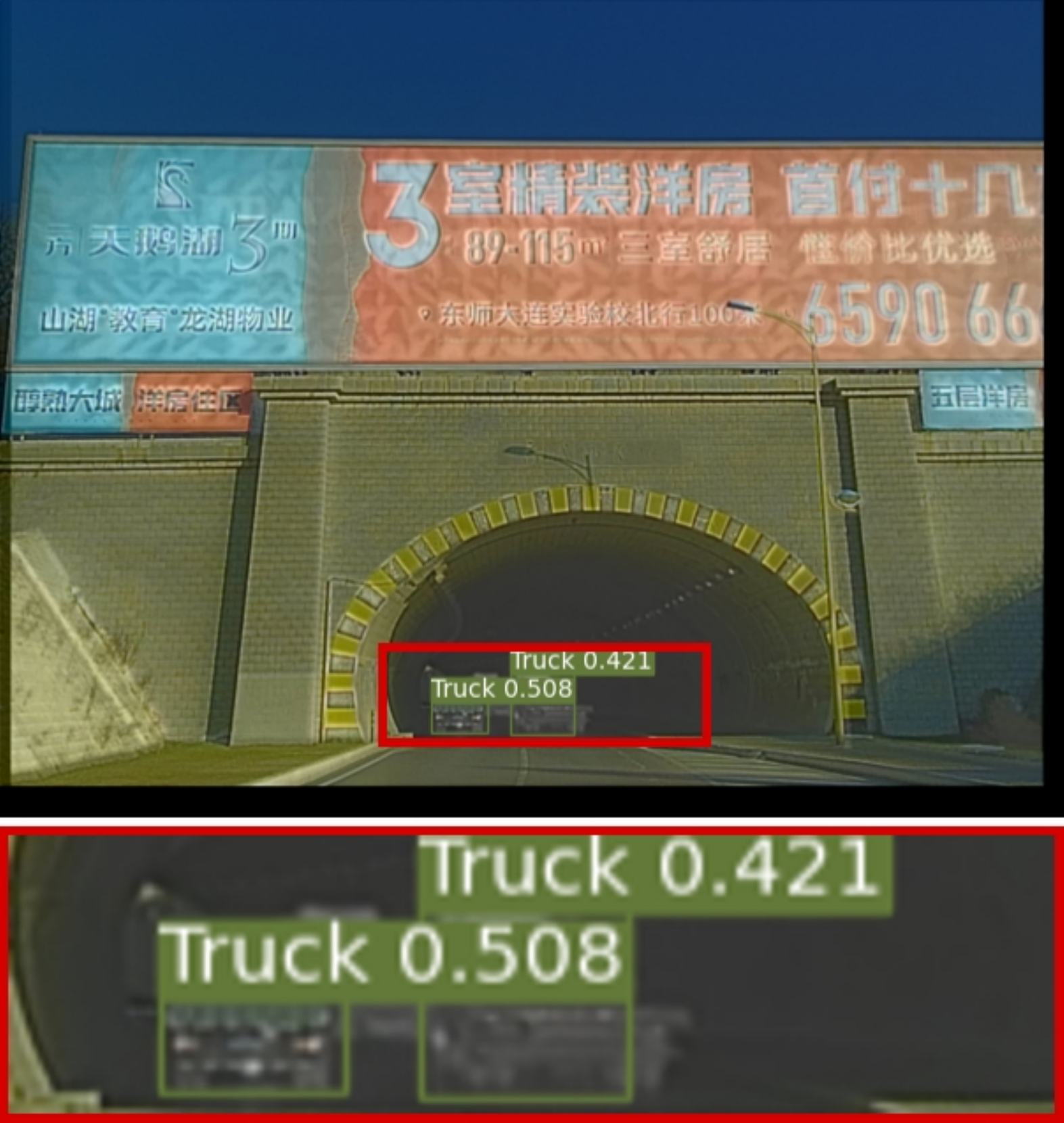}
		&\includegraphics[width=0.14\textwidth]{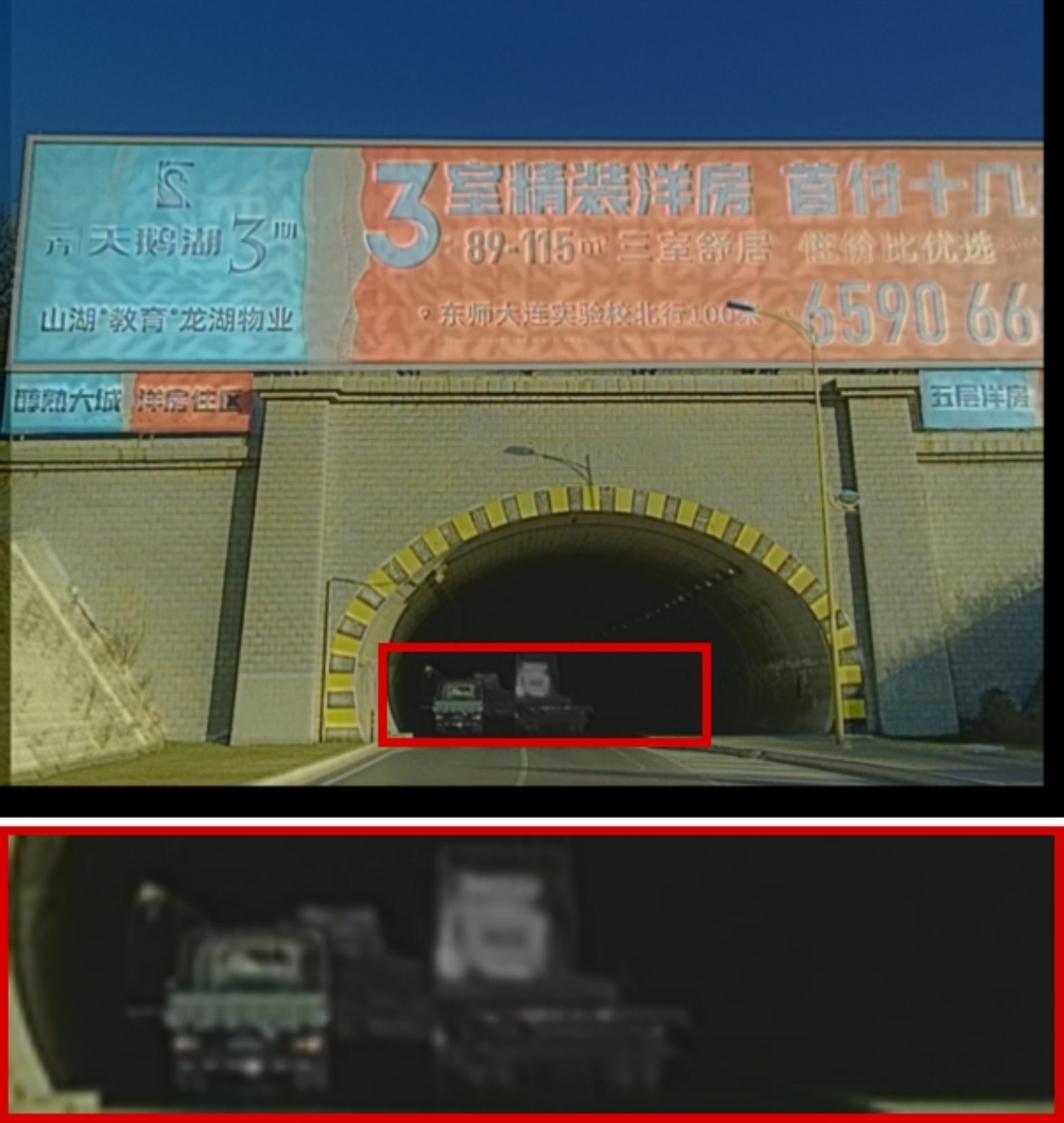}
		&\includegraphics[width=0.14\textwidth]{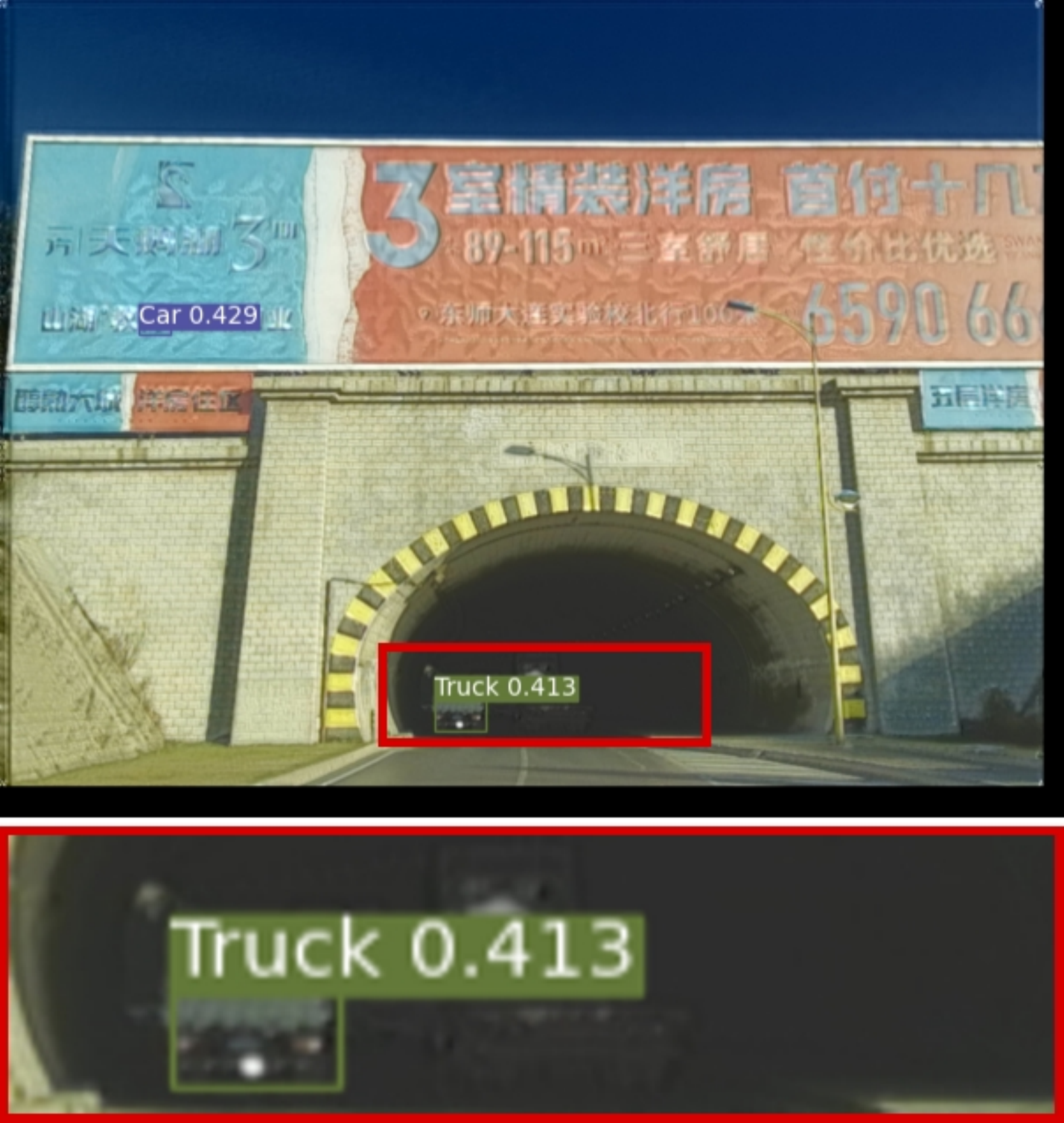}
		&\includegraphics[width=0.14\textwidth]{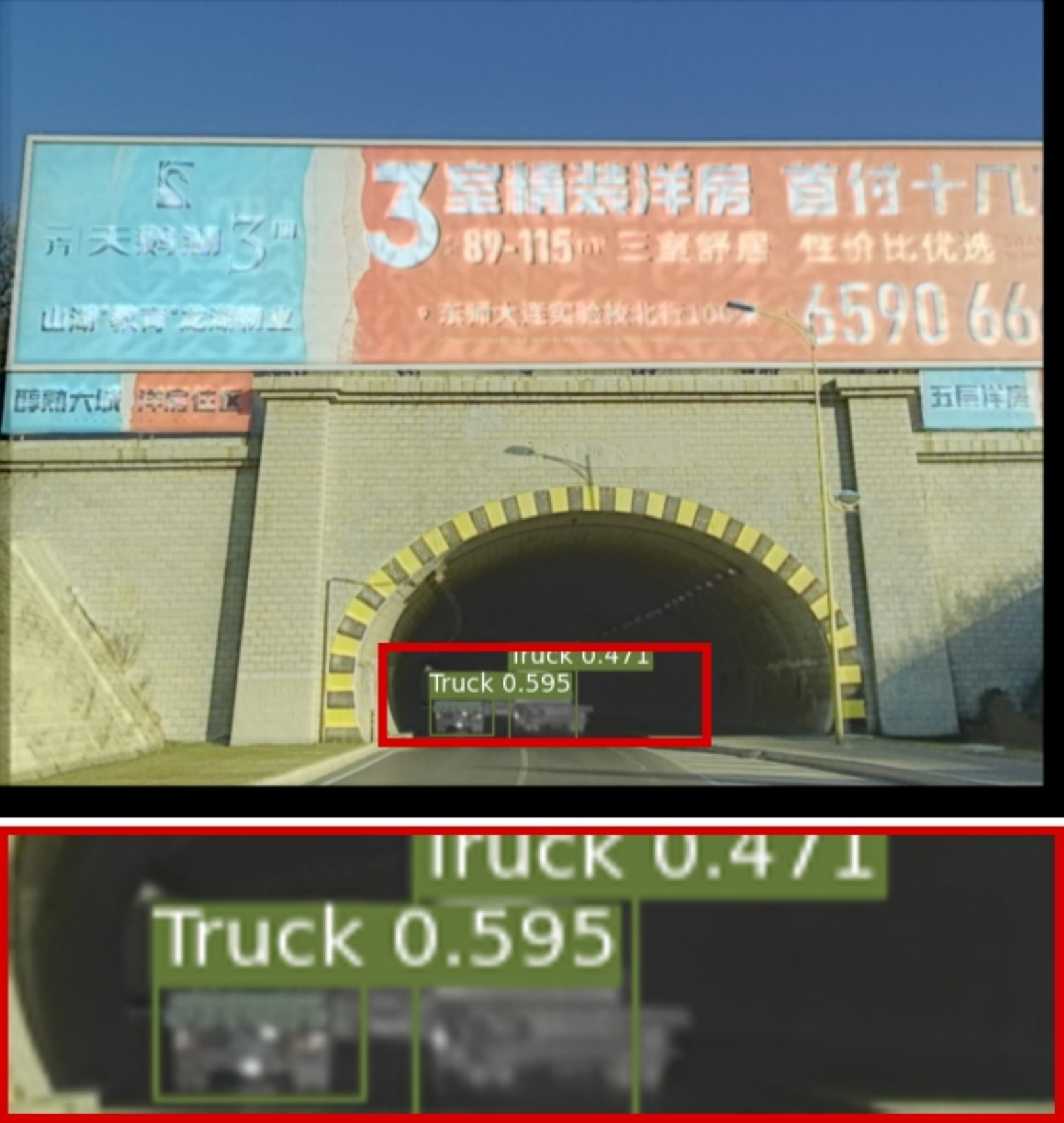}\\
		\includegraphics[width=0.14\textwidth]{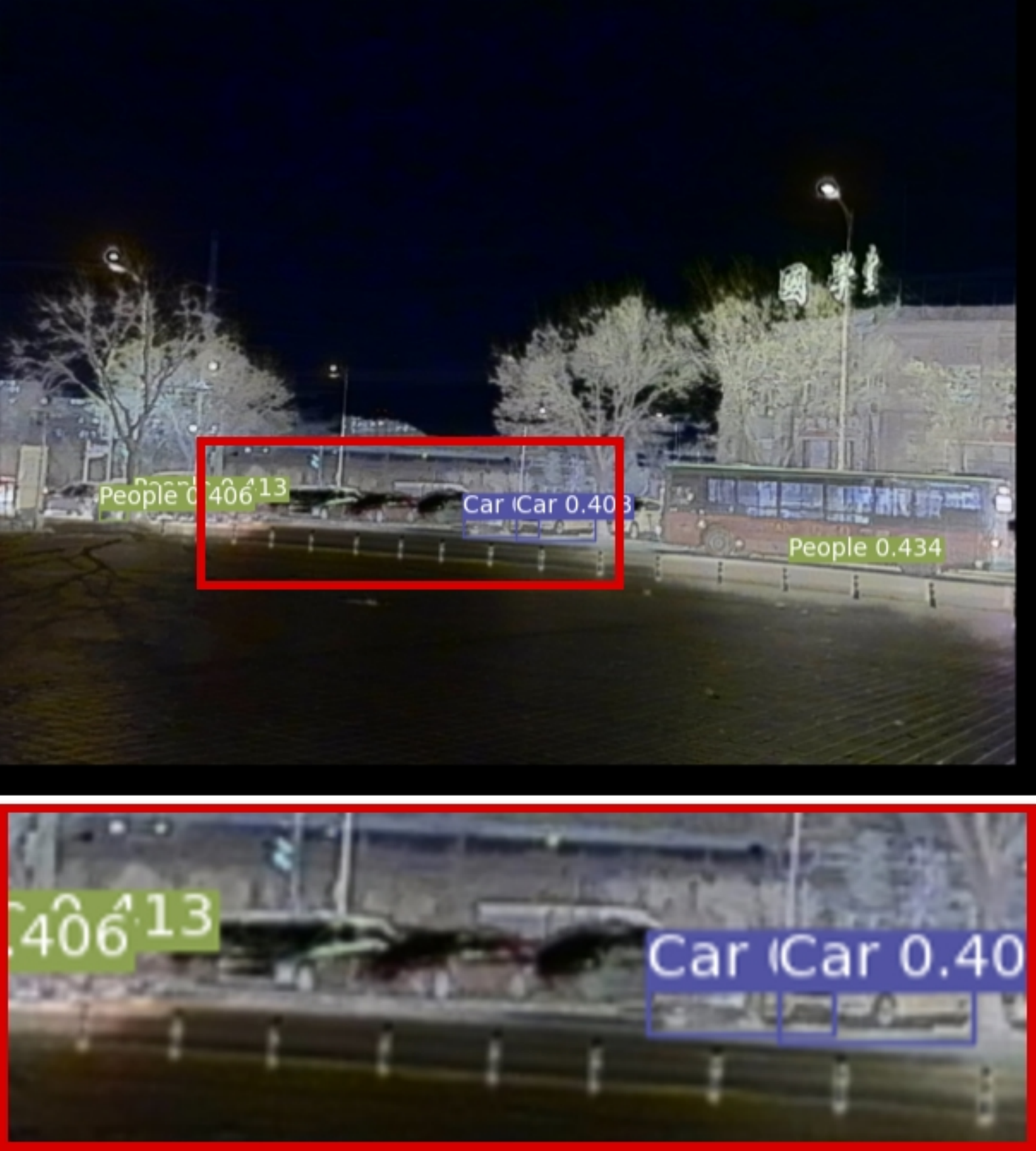}
		&\includegraphics[width=0.14\textwidth]{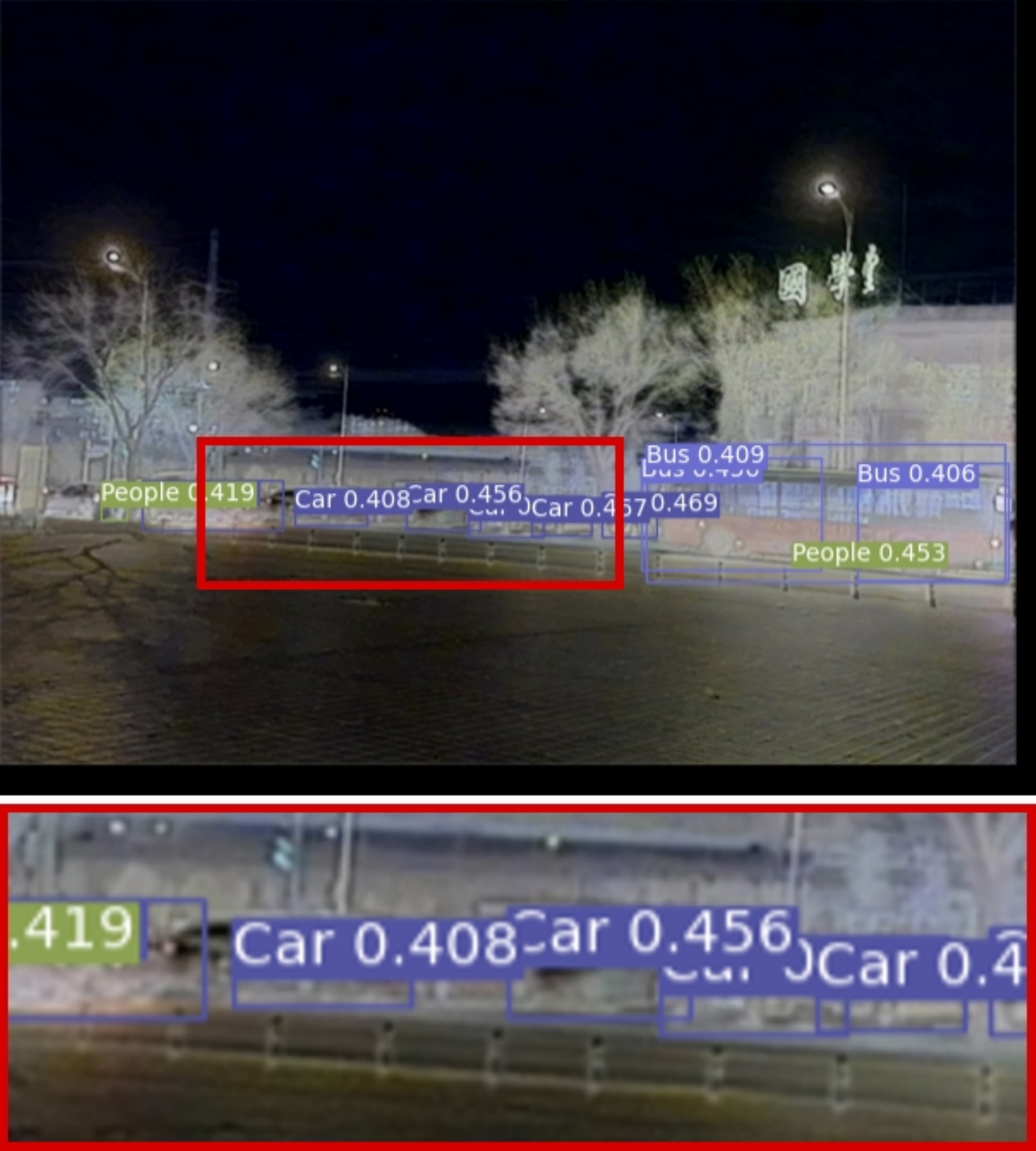}
		&\includegraphics[width=0.14\textwidth]{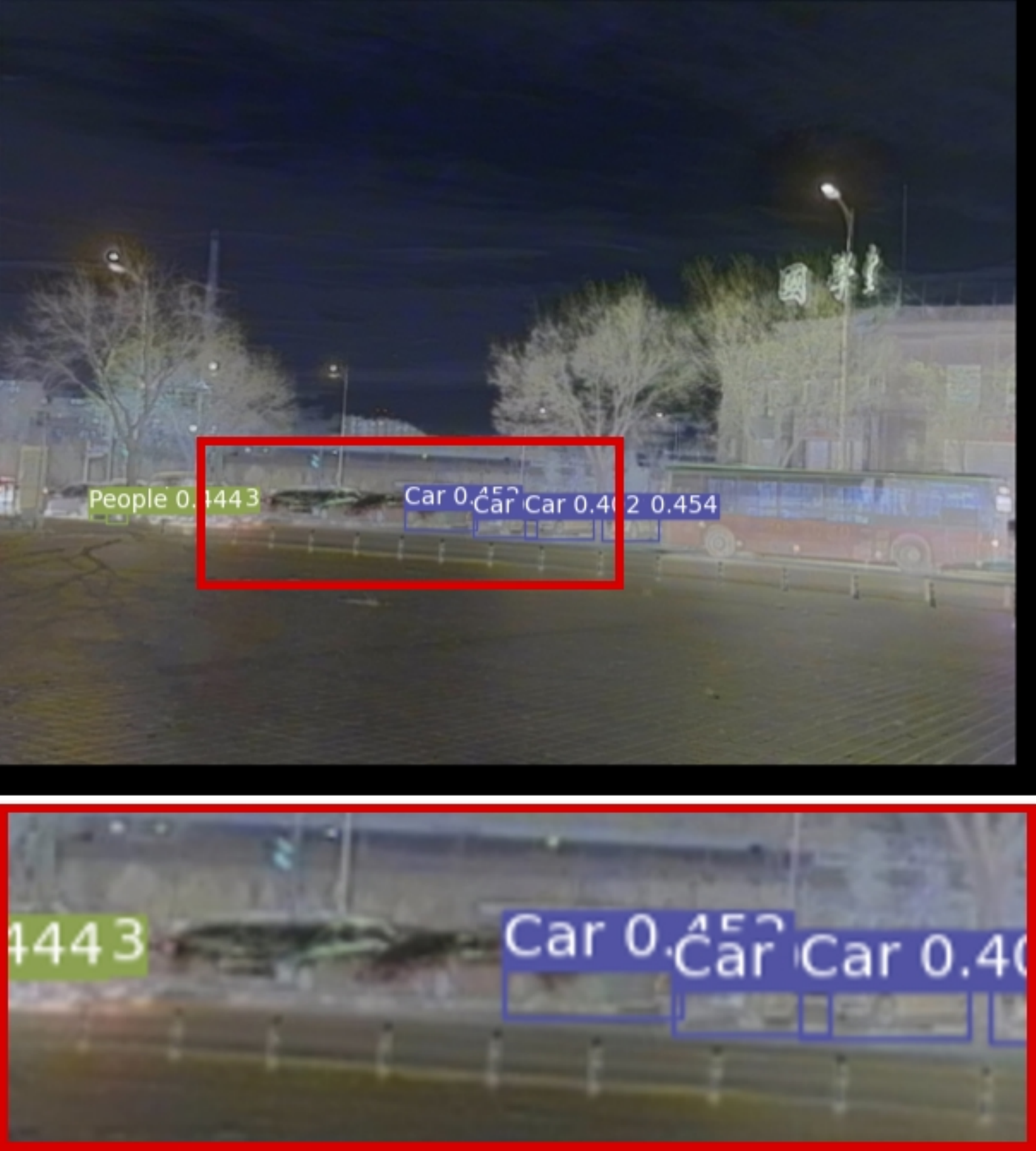}
		&\includegraphics[width=0.14\textwidth]{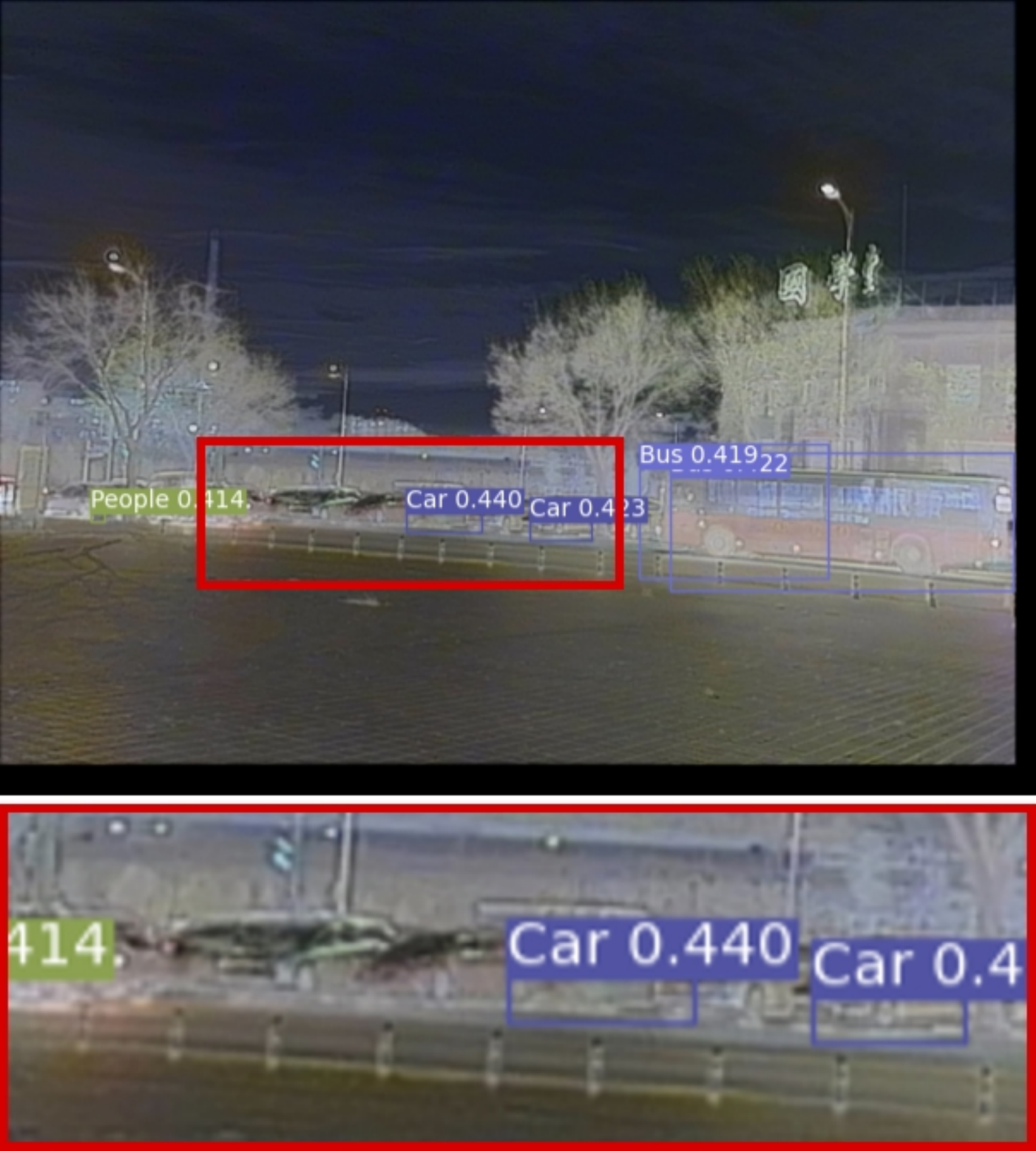}
		&\includegraphics[width=0.14\textwidth]{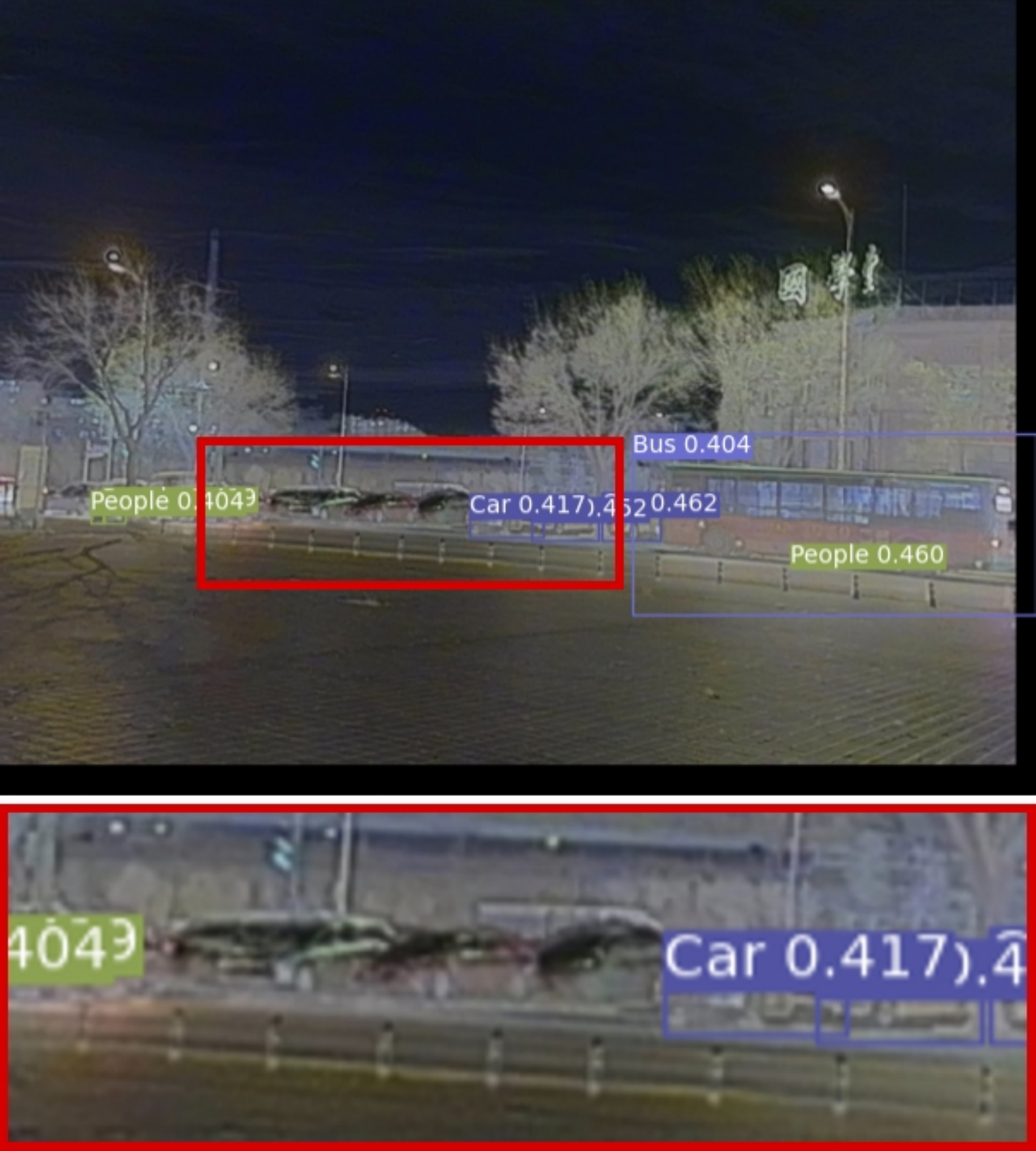}
		&\includegraphics[width=0.14\textwidth]{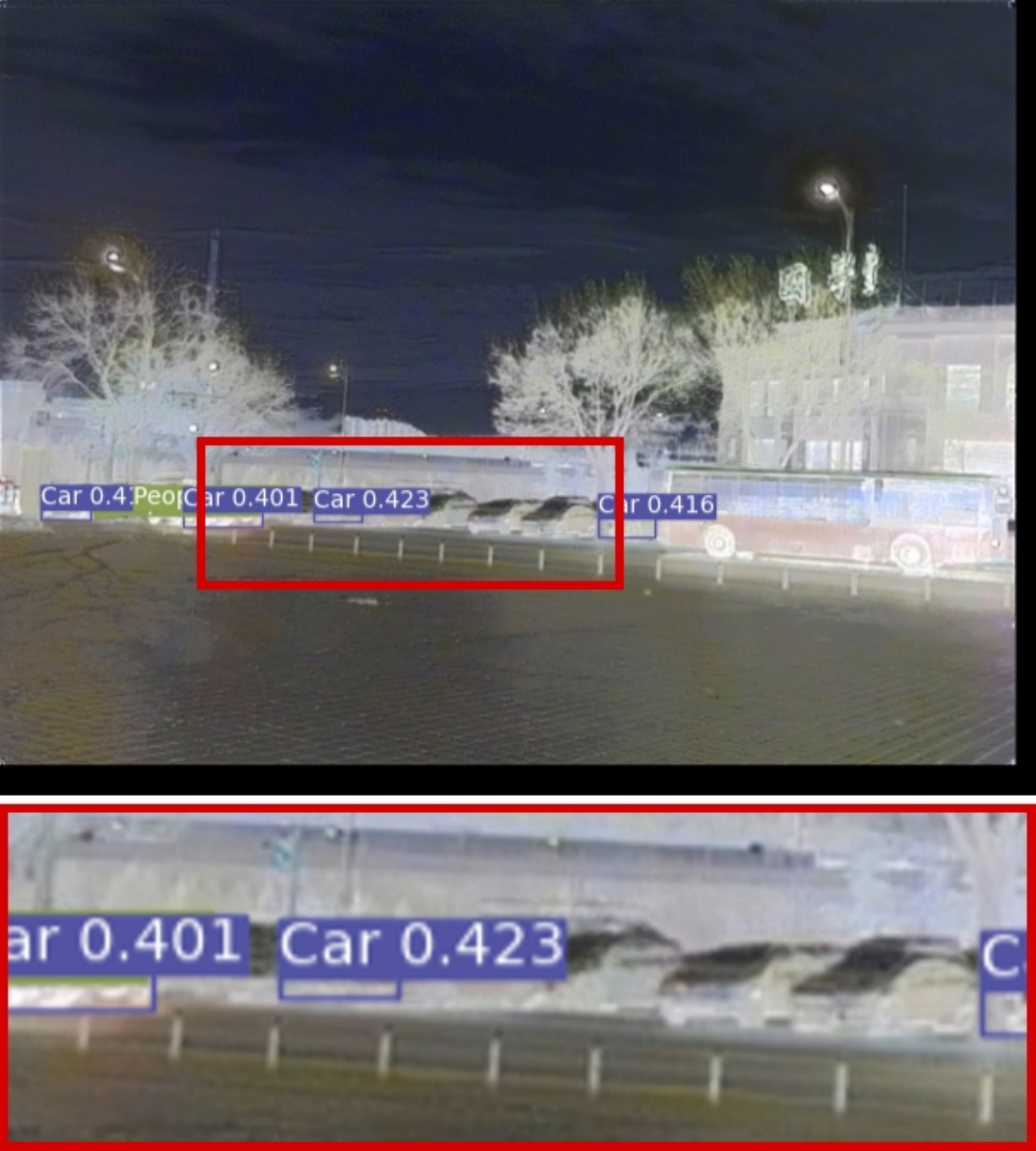}
		&\includegraphics[width=0.14\textwidth]{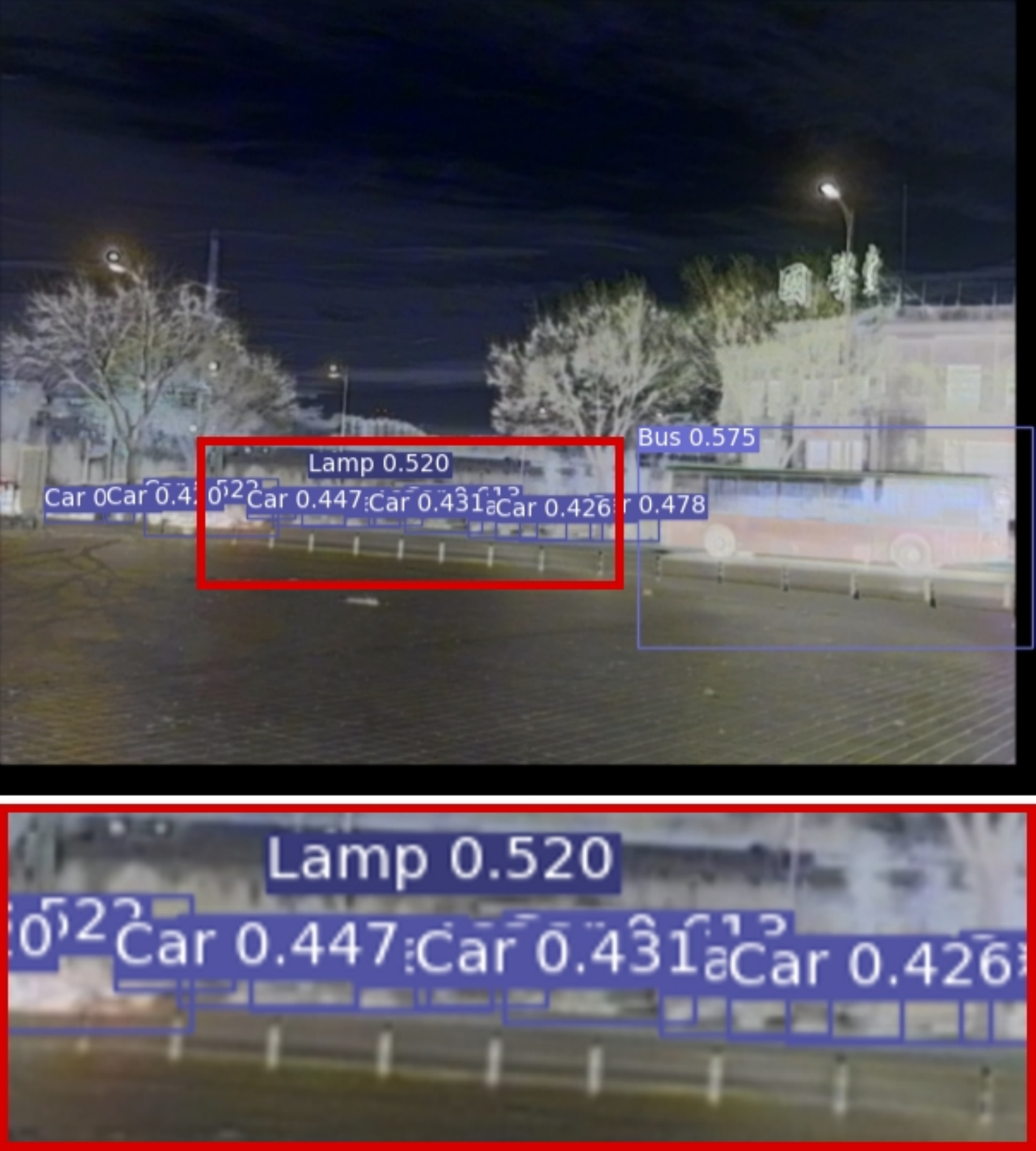}\\
		
		\footnotesize	DIDFuse &\footnotesize ReCoNet&\footnotesize UMFusion&\footnotesize SDNet&\footnotesize U2Fusion&\footnotesize TarDAL& \footnotesize Ours
		\\
	\end{tabular}

	\caption{Detection comparison with advanced approaches under two difficult scenarios (\emph{i.e.,} tunnel and  dense objects in the darkness).}
	\label{fig:detection}
\end{figure*}
\begin{table*}[htb]
	\centering
	\footnotesize
	\renewcommand{\arraystretch}{1.1}
	\setlength{\tabcolsep}{0.6mm}{
		\begin{tabular}{c|ccccccc|ccccc c|ccc}
			\hline
			\multirow{2}{*}{Methods} & \multicolumn{7}{c|}{M3FD}                                                                                                                          &
			\multicolumn{6}{c|}{Multi-Spectral}                                                                          & \multicolumn{3}{c}{Efficiency Analysis}                 \\ \cline{2-17} 
			& \cellcolor{gray!20} {Lamp} &\cellcolor{gray!20} {Car} & \cellcolor{gray!20}{Bus} & \cellcolor{gray!20}{Motor} & \cellcolor{gray!20}{Truck} & \cellcolor{gray!20}{People} & \cellcolor{gray!20} mAP $\uparrow$ & \multicolumn{1}{c}{\cellcolor{gray!20}{CStop}} & \cellcolor{gray!20}{Car} & \cellcolor{gray!20}{Person}  & \cellcolor{gray!20}{Bump}   & \cellcolor{gray!20}{Bike} & \cellcolor{gray!20}{mAP} $\uparrow$ &  \cellcolor{gray!20}{Size(M)$\downarrow$}& \cellcolor{gray!20}{FLOPs(G)$\downarrow$}& \cellcolor{gray!20}{Time(S)$\downarrow$}\\ \hline\hline
			DDcGAN	 & \multicolumn{1}{c|}{0.247} & \multicolumn{1}{c|}{0.664}  & \multicolumn{1}{c|}{0.451}   & \multicolumn{1}{c|}{0.312}   & \multicolumn{1}{c|}{0.355} & \multicolumn{1}{c|}{0.444} & 0.412  & \multicolumn{1}{c|}{0.312} & \multicolumn{1}{c|}{0.360} & \multicolumn{1}{c|}{0.199} & \multicolumn{1}{c|}{0.167}  & \multicolumn{1}{c|}{0.347} & \multicolumn{1}{c|}{0.277} & \multicolumn{1}{c|}{1.097}& \multicolumn{1}{c|}{896.84}& \multicolumn{1}{c}{0.211}\\ \hline
			DenseFuse	 & \multicolumn{1}{c|}{0.249} & \multicolumn{1}{c|}{0.694}  & \multicolumn{1}{c|}{0.432}   & \multicolumn{1}{c|}{0.319}   & \multicolumn{1}{c|}{0.346} & \multicolumn{1}{c|}{0.501} & 0.424  & \multicolumn{1}{c|}{\textcolor{blue}{\textbf{0.425}}} & \multicolumn{1}{c|}{0.515} & \multicolumn{1}{c|}{\textcolor{blue}{\textbf{0.541}}} & \multicolumn{1}{c|}{0.377}  & \multicolumn{1}{c|}{0.409} & \multicolumn{1}{c|}{0.453} & \multicolumn{1}{c|}{\textcolor{blue}{\textbf{0.074}}}& \multicolumn{1}{c|}{48.96}& \multicolumn{1}{c}{0.251}\\ \hline
			AUIF	 & \multicolumn{1}{c|}{0.244} & \multicolumn{1}{c|}{0.693}  & \multicolumn{1}{c|}{0.406}   & \multicolumn{1}{c|}{0.282}   & \multicolumn{1}{c|}{0.326} & \multicolumn{1}{c|}{0.496} & 0.408  & \multicolumn{1}{c|}{0.387} & \multicolumn{1}{c|}{0.528} & \multicolumn{1}{c|}{0.479} & \multicolumn{1}{c|}{0.222}  & \multicolumn{1}{c|}{0.424} & \multicolumn{1}{c|}{0.408} & \multicolumn{1}{c|}{\textcolor{red}{\textbf{0.012}}}&  \multicolumn{1}{c|}{\textcolor{red}{\textbf{0.014}}} & \multicolumn{1}{c}{0.166}\\ \hline
			DIDFuse	 & \multicolumn{1}{c|}{0.300} & \multicolumn{1}{c|}{0.722}  & \multicolumn{1}{c|}{0.373}   & \multicolumn{1}{c|}{0.305}   & \multicolumn{1}{c|}{0.444} & \multicolumn{1}{c|}{0.536} & 0.466  & \multicolumn{1}{c|}{0.368} & \multicolumn{1}{c|}{0.490} & \multicolumn{1}{c|}{0.504} & \multicolumn{1}{c|}{0.277}  & \multicolumn{1}{c|}{0.373} & \multicolumn{1}{c|}{0.402} & \multicolumn{1}{c|}{0.373}& \multicolumn{1}{c|}{103.56}& \multicolumn{1}{c}{0.118}\\ \hline
			MEFIF	 & \multicolumn{1}{c|}{0.285} & \multicolumn{1}{c|}{0.719}  & \multicolumn{1}{c|}{0.442}   & \multicolumn{1}{c|}{0.321}   & \multicolumn{1}{c|}{0.418} & \multicolumn{1}{c|}{0.540} & 0.454  & \multicolumn{1}{c|}{\textcolor{red}{\textbf{0.432}}} & \multicolumn{1}{c|}{\textcolor{blue}{\textbf{0.531}}} & \multicolumn{1}{c|}{0.539} & \multicolumn{1}{c|}{\textcolor{blue}{\textbf{0.377}}} & \multicolumn{1}{c|}{0.419} & \multicolumn{1}{c|}{\textcolor{blue}{\textbf{0.460}}} & \multicolumn{1}{c|}{0.705}& \multicolumn{1}{c|}{48.92}& \multicolumn{1}{c}{0.141}\\ \hline
			
			ReCoNet	 & \multicolumn{1}{c|}{\textcolor{blue}{\textbf{0.336}}} & \multicolumn{1}{c|}{\textcolor{blue}{\textbf{0.744}}}  & \multicolumn{1}{c|}{\textcolor{blue}{\textbf{0.482}}}   & \multicolumn{1}{c|}{0.311}   & \multicolumn{1}{c|}{\textcolor{blue}{\textbf{0.445}}} & \multicolumn{1}{c|}{\textcolor{blue}{\textbf{0.556}}} & \textcolor{blue}{\textbf{0.479}}  & \multicolumn{1}{c|}{0.401} & \multicolumn{1}{c|}{0.507} & \multicolumn{1}{c|}{0.401} & \multicolumn{1}{c|}{0.377}  & \multicolumn{1}{c|}{0.397} & \multicolumn{1}{c|}{0.417} & \multicolumn{1}{c|}{0.209}& \multicolumn{1}{c|}{\textcolor{blue}{\textbf{12.54}}}& \multicolumn{1}{c}{0.051}\\ \hline

			UMFusion	 & \multicolumn{1}{c|}{0.254} & \multicolumn{1}{c|}{0.692}  & \multicolumn{1}{c|}{0.424}   & \multicolumn{1}{c|}{0.309}   & \multicolumn{1}{c|}{0.343} & \multicolumn{1}{c|}{0.518} & 0.423  & \multicolumn{1}{c|}{0.389} & \multicolumn{1}{c|}{0.525} & \multicolumn{1}{c|}{0.504} & \multicolumn{1}{c|}{0.361}  & \multicolumn{1}{c|}{0.439} & \multicolumn{1}{c|}{0.442} & \multicolumn{1}{c|}{0.629}& \multicolumn{1}{c|}{174.69}& \multicolumn{1}{c}{0.044}\\ \hline

			SDNet	 & \multicolumn{1}{c|}{0.265} & \multicolumn{1}{c|}{0.702}  & \multicolumn{1}{c|}{0.429}   & \multicolumn{1}{c|}{0.344}   & \multicolumn{1}{c|}{0.331} & \multicolumn{1}{c|}{0.515} & 0.431  & \multicolumn{1}{c|}{0.359} & \multicolumn{1}{c|}{0.503} & \multicolumn{1}{c|}{0.495} & \multicolumn{1}{c|}{0.167}  & \multicolumn{1}{c|}{0.433} & \multicolumn{1}{c|}{0.391} & \multicolumn{1}{c|}{0.067}& \multicolumn{1}{c|}{37.35}& \multicolumn{1}{c}{0.045}\\ \hline

			U2Fusion	 & \multicolumn{1}{c|}{0.312} & \multicolumn{1}{c|}{0.724}  & \multicolumn{1}{c|}{0.475}   & \multicolumn{1}{c|}{\textcolor{blue}{\textbf{0.352}}}   & \multicolumn{1}{c|}{0.392} & \multicolumn{1}{c|}{0.534} & 0.465  & \multicolumn{1}{c|}{0.419} & \multicolumn{1}{c|}{0.542} & \multicolumn{1}{c|}{0.501} & \multicolumn{1}{c|}{0.333}  & \multicolumn{1}{c|}{\textcolor{blue}{\textbf{0.453}}} & \multicolumn{1}{c|}{0.449} & \multicolumn{1}{c|}{0.659}& \multicolumn{1}{c|}{366.34}& \multicolumn{1}{c}{0.123}\\ \hline
			TarDAL	 & \multicolumn{1}{c|}{0.229} & \multicolumn{1}{c|}{0.652}  & \multicolumn{1}{c|}{0.425}   & \multicolumn{1}{c|}{0.285}   & \multicolumn{1}{c|}{0.317} & \multicolumn{1}{c|}{0.515} & 0.404  & \multicolumn{1}{c|}{0.343} & \multicolumn{1}{c|}{0.495} & \multicolumn{1}{c|}{0.513} & \multicolumn{1}{c|}{0.167}  & \multicolumn{1}{c|}{0.392} & \multicolumn{1}{c|}{0.382} & \multicolumn{1}{c|}{0.297}& \multicolumn{1}{c|}{82.37}& \multicolumn{1}{c}{\textcolor{red}{\textbf{0.001}}} \\ \hline
			
			Ours	 & \multicolumn{1}{c|}{\textcolor{red}{\textbf{0.351}}} & \multicolumn{1}{c|}{\textcolor{red}{\textbf{0.759}}}  & \multicolumn{1}{c|}{\textcolor{red}{\textbf{0.511}}}   & \multicolumn{1}{c|}{\textcolor{red}{\textbf{0.413}}}   & \multicolumn{1}{c|}{\textcolor{red}{\textbf{0.551}}} & \multicolumn{1}{c|}{\textcolor{red}{\textbf{0.598}}} & \textcolor{red}{\textbf{0.531}}  & \multicolumn{1}{c|}{0.423} & \multicolumn{1}{c|}{\textcolor{red}{\textbf{0.586}}} & \multicolumn{1}{c|}{\textcolor{red}{\textbf{0.574}}} & \multicolumn{1}{c|}{\textcolor{red}{\textbf{0.479}}}  & \multicolumn{1}{c|}{\textcolor{red}{\textbf{0.525}}} & \multicolumn{1}{c|}{\textcolor{red}{\textbf{0.504}}} & \multicolumn{1}{c|}{0.417}& \multicolumn{1}{c|}{106.02}& \multicolumn{1}{c}{\textcolor{red}{\textbf{0.001}}} \\ \hline\hline
			
	\end{tabular}}

	\caption{ Quantitative  results of object detection  on the M3FD and Multi-Spectral datasets and efficiency analysis. The best result is in red whereas the second best one is in blue.}~\label{tab:detec}
\end{table*}
\subsection{Loss Functions}
In this part, we will elaborate the concrete loss functions to define $\phi$ and $\Phi$ respectively, which can be divided into two parts for visual quality and semantic perception respectively.

As for the learning of discriminator $\mathcal{T}_\mathtt{V}$~\cite{isola2017image}, we introduce the generative adversarial mechanism to  discriminate the visual quality of image fusion. In concrete, we first construct the pseudo fused images $\mathbf{u}_{m}$ to 
maintain the salient informations by saliency weight maps ($\mathtt{m}_{1}$ and $\mathtt{m}_{2}$) using VSM~\cite{ma20171123}, \emph{i.e.,}  $\mathbf{u}_{m} = \mathtt{m}_{1}\mathbf{x} +\mathtt{m}_{2}\mathbf{y}$. We also  gradient-penalty Wasserstein strategy~\cite{gulrajani2017improved} to guarantee the learning stability, thus the concrete formulation to train $\mathcal{T}_\mathtt{v}$ can be written as
\begin{equation}\label{eq:gangp}
\phi_\mathtt{v} = \mathbb{E}_{\tilde{\mathbf{s}}\sim\mathbb{P}_\mathtt{fake}}\mathcal{T}_\mathtt{V}(\mathbf{u}) - \mathbb{E}_{{\mathbf{s}}\sim\mathbb{P}_\mathtt{real}}\mathcal{T}_\mathtt{V}(\mathbf{u}_{m})\\+\eta \mathtt{R}_\mathtt{Penalty},
\end{equation} 
where  $ \mathtt{R}_\mathtt{Penalty}$ is the penalty term, calculated by $ \mathbb{E}_{\tilde{\mathbf{s}}\sim\mathbb{P}_\mathtt{fake}} [(\|\nabla_{\mathbf{u}}\mathcal{T}_\mathtt{V}(\mathbf{u})\|_{2}-1)^2]$ and $\eta$ is a trade-off term.

Furthermore, image fusion network $\mathcal{F}$ can be considered as the generator. In order to balance the pixel intensity and avoid the texture artifact, we also leverage the pixel error loss for the fusion learning, \emph{i.e.,} 
\begin{equation}\label{eq:gangp}
\Phi_\mathtt{v} =\lVert \mathbf{u}- \mathtt{m}_{1}\mathbf{x}\rVert^2_2 + \lVert \mathbf{u}- \mathtt{m}_{2}\mathbf{y}\rVert^2_2 - \mathbb{E}_{{\mathbf{s}}\sim\mathbb{P}_\mathtt{real}}\mathcal{T}_\mathtt{V}(\mathbf{u}).
\end{equation} 

As for the semantic perception optimization of $\mathcal{T}_\mathtt{P}$, we adopt the common task-specific loss functions to training the perception-related objectives (\emph{i.e.,}  $\Phi_\mathtt{P}$ and $\phi_\mathtt{P}$).
For object detection, we utilize the hybrid loss functions from FCOS~\cite{tian2019fcos} to define the objective. As for the semantic segmentation, common to previous literature, we utilize the cross-entropy loss function.



%

\section{Experiments}
\begin{figure*}[!htb]
	\centering
	\setlength{\tabcolsep}{1pt} 
	
	\includegraphics[width=0.99\textwidth,height=0.22\textheight]{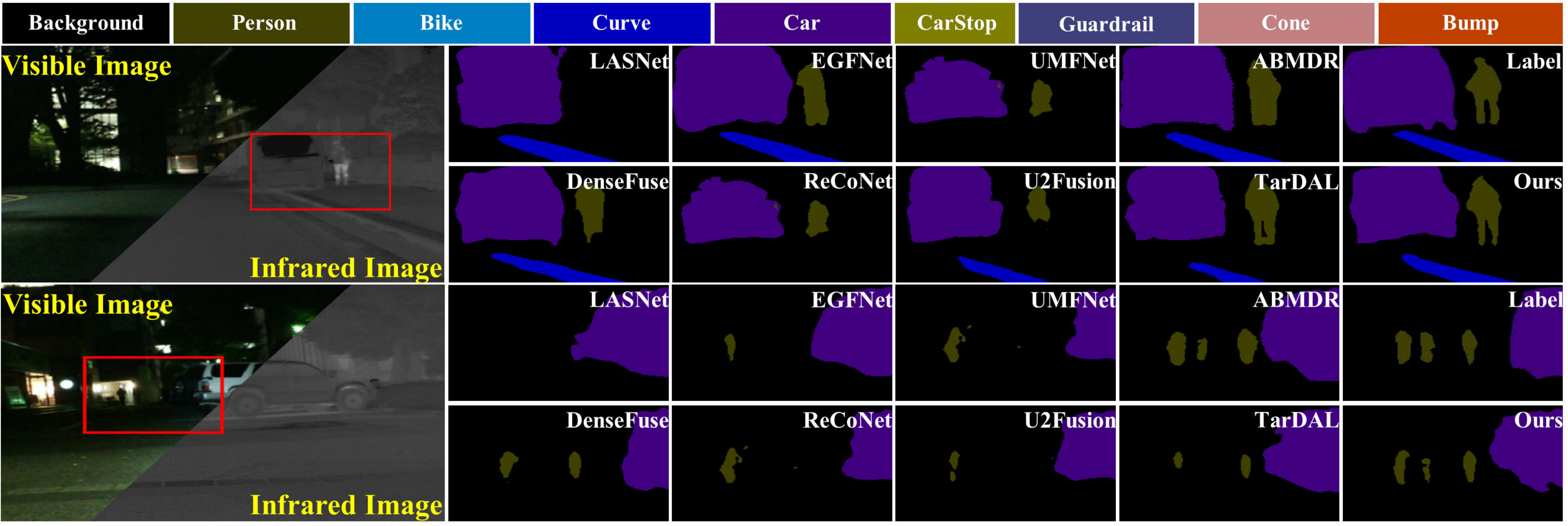}
	
	\caption{Qualitative comparisons with eight advanced competitors under nighttime on the MFNet benchmark.}
	\label{fig:MSPresult}
\end{figure*}

\begin{table*}[htb]
	\centering
	\footnotesize
	\renewcommand{\arraystretch}{1.1}
	\setlength{\tabcolsep}{0.9mm}{
		\begin{tabular}{c|cc|cc|cc|cc|cc|cc|cc|cc|cc}
			\hline
			\multirow{2}{*}{Methods} &			\multicolumn{2}{c}{Unlabel}&  \multicolumn{2}{c}{Car} & \multicolumn{2}{c}{Person} & \multicolumn{2}{c}{Bike}& \multicolumn{2}{c}{Curve}  & \multicolumn{2}{c}{Car Stop}&
			\multicolumn{2}{c}{Cone}& \multicolumn{2}{c|}{Bump}
			&\multirow{2}{*}{ mAcc$\uparrow$}&\multirow{2}{*}{mIoU$\uparrow$} \\ \cline{2-17} 
			&\cellcolor{gray!20}{Acc$\uparrow$} &\cellcolor{gray!20}{IoU$\uparrow$}  &\cellcolor{gray!20}{Acc$\uparrow$} &\cellcolor{gray!20}{IoU$\uparrow$} &\cellcolor{gray!20}{Acc$\uparrow$} &\cellcolor{gray!20}{IoU$\uparrow$} &\cellcolor{gray!20}{Acc$\uparrow$} &\cellcolor{gray!20}{IoU$\uparrow$}  &\cellcolor{gray!20}{Acc$\uparrow$} &\cellcolor{gray!20}{IoU$\uparrow$}   &\cellcolor{gray!20}{Acc$\uparrow$} &\cellcolor{gray!20}{IoU$\uparrow$}  &\cellcolor{gray!20}{Acc$\uparrow$} &\cellcolor{gray!20}{IoU$\uparrow$} &\cellcolor{gray!20}{Acc$\uparrow$} &\cellcolor{gray!20}{IoU$\uparrow$}   &\multicolumn{1}{c}{} \\
			\hline\hline
			
			\cellcolor{blue!15}LASNet &99.2 &97.4  &94.9  &84.2 & 81.7 &67.1 & 82.1 &	56.9 & {\textcolor{blue}{\textbf{70.7}}} &41.1 & 56.8 &{\textcolor{red}{\textbf{39.6}}}& 58.1 &{\textcolor{blue}{\textbf{48.8}}}& 77.2&	40.1  &  {\textcolor{blue}{\textbf{75.4}}} &{\textcolor{blue}{\textbf{54.9}}}\\\hline
			\cellcolor{blue!15}EGFNet & {\textcolor{red}{\textbf{99.3}}}  &97.7 &{{{95.7}}}   &{\textcolor{red}{\textbf{87.6}}} & {\textcolor{blue}{\textbf{89.0}}}& 69.8 &80.6 &	58.8 &  {\textcolor{red}{\textbf{71.5}}} &{\textcolor{blue}{\textbf{42.8}}} & 48.7 & 33.8 & 65.3& 48.3 &71.1& 47.1&	72.7 & 54.8\\			\hline
			\cellcolor{blue!15} ABMDR &  {\textcolor{red}{\textbf{99.3}}}  & {\textcolor{red}{\textbf{98.4}}}  &94.3  &	84.8 & {\textcolor{red}{\textbf{90.0}}} &69.6 &75.7 &	60.3 & 64.0 &{\textcolor{red}{\textbf{45.1}}} & 44.1 &	33.1& 61.7 &	47.4 & 66.2 &	 {\textcolor{blue}{\textbf{50.0}}}   & 69.5 &	54.8 \\
			\hline
			DenseFuse & 98.4& 97.8& 93.5& 82.8& 86.0& 67.7& 81.6 &60.2& 66.5 &42.5 &61.4& 12.6& 80.5& 39.1& 67.6& 44.1& 71.1& 50.1\\\hline
			AUIF&97.9 &97.4& 95.1& 80.2& 80.8& 54.0& 78.4& 57.0& 68.8& 27.0& 62.2& 30.1& 81.9& 40.0 &{\textcolor{red}{\textbf{96.5}}}  &24.8& 73.8& 45.8\\ \hline
			DIDFuse & 98.1&	96.9&89.1&78.0& 82.9&	59.6& 70.5&	53.4	& 34.2&20.6&49.0&	25.1& 82.4&34.7&59.5& 20.0& 63.1&	43.3	 \\\hline
			ReCoNet & 97.9	&97.2&\textcolor{red}{\textbf{96.0}}&79.3& 85.4&	60.9& 77.5	&58.2	& 43.2&22.6	&56.8&32.8& 73.5	&36.7	& 36.2&13.5& 63.0&	44.6      \\\hline
			UMFusion   &98.1 &97.5& 95.4& 81.4 &88.7 &62.5& 78.5& 60.1 &47.8& 25.5& 61.2 &26.2 &67.6& 39.6& 85.6 &46.8& 69.6 &49.1\\ \hline
			SDNet &98.4& 97.7&96.1&	83.1& 85.9& {\textcolor{blue}{\textbf{70.5}}}& 75.6 &60.2 & 60.9& 35.6 &  {\textcolor{red}{\textbf{76.6}}}& 28.1 &   {\textcolor{red}{\textbf{88.6}}}&46.1& 71.1 &45.5& 73.7 &52.8 \\ \hline
			U2Fusion& 98.3&	97.7&94.3  &	82.7 & 87.0 &64.1 & 78.9 &	\textcolor{blue}{\textbf{61.0}} & 56.2 &	35.5 & {\textcolor{blue}{\textbf{75.2}}} &	24.0 & 80.5 &46.3 &  81.3 &49.9  & 72.6 &	51.4 \\\hline
			TarDAL   &98.4& 97.5&89.2 &	79.5& 87.1& 67.3& 77.0 &	59.9 & 53.3& 29.1 &  72.5& 22.0  &  84.2&35.6& 70.6 &40.4& 70.6 &48.2 \\ \hline
			SeaFusion   &98.2& 97.7& 94.4& 82.2 &88.1& 67.7 &{\textcolor{blue}{\textbf{83.9}}}& 59.7 &65.7& 37.3& 72.4& 16.8& {\textcolor{blue}{\textbf{87.3}}} &41.3 &{\textcolor{blue}{\textbf{85.7}}} & 27.2& {\textcolor{red}{\textbf{75.5}}}& 48.1\\ \hline
			{Ours}  &98.7&  {\textcolor{blue}{\textbf{98.0}}} & {\textcolor{blue}{\textbf{95.7}}}   &	{\textcolor{blue}{\textbf{86.2}}}  &  84.4& {\textcolor{red}{\textbf{71.6}}} &  {\textcolor{red}{\textbf{84.4}}} &	 {\textcolor{red}{\textbf{63.6}}} & 60.4 &	 40.7& 69.1 & \textcolor{blue}{\textbf{37.3}} &  {\textcolor{blue}{\textbf{87.3}}} & {\textcolor{red}{\textbf{49.9}}} &  83.0  &{\textcolor{red}{\textbf{51.2}}}  &  73.6 &	{\textcolor{red}{\textbf{55.4}}}  \\
			\hline\hline
	\end{tabular} }	
	\caption{ Quantitative results of  semantic segmentation with different methods on the {MFNet} dataset. }~\label{tab:seg}
\end{table*}

\subsection{Implementation Details}
We implemented the experiments on five representative datasets (TNO, RoadScene~\cite{U2Fusion2020} and M3FD
for visual evaluations, M3FD~\cite{TarDAL} and Multi-Spectral~\cite{takumi2017multispectral} for detection, and MFNet~\cite{ha2017mfnet} for segmentation). SGD optimizer is utilized to update the parameters of each module. As for the perception task, the initialized learning rate is $2e^{-4}$ and will be decreased to $2e^{-6}$ with a multi-step decay strategy.
As for the optimization of fusion, we utilize the same learning rate. 
All experiments are implemented with the PyTorch framework and on an NVIDIA Tesla V100 GPU.

\subsection{Evaluation in Multi-modality Image Fusion}

We conducted qualitative and quantitative analyses with ten state-of-the-art competitors, including DDcGAN~\cite{xu2019learning}, AUIF~\cite{zhao2021efficient}, MFEIF~\cite{MFEIF2021}, SDNet~\cite{zhang2021sdnet}, DIDFuse~\cite{zhao2020didfuse}, DenseFuse~\cite{li2018densefuse}, ReCoNet~\cite{reconet}, UMFusion~\cite{UMFusion}, TarDAL~\cite{TarDAL} and U2Fusion~\cite{U2Fusion2020}. \footnote{  ``Fusion with detection'' is selected for  comparison.} 

\paragraph{Qualitative comparisons.}
The qualitative results on diverse challenging scenarios are depicted in Figure~\ref{fig:contristive}. It can be clearly observed that our method outperforms other methods in two eye-catching aspects. First, our method sufficiently reserves the high-contrast information from infrared images. For instance, in all scenarios, the buildings and pedestrians are best highlighted. Furthermore, our method  reserves visible structure on the premise of retaining significant infrared characteristics. Overall, our method highlights the vital information (\emph{e.g.,}  pedestrians) and suppresses the disturbing information (\emph{e.g.,}  large darkness).

\paragraph{Quantitative comparisons.} 
We also report the numerical results with the other ten fusion competitors on TNO, RoadScene, and M3FD in Table~\ref{tab:fusion}. We adopt three objective metrics for analysis.  We can clearly observe that our method demonstrates superiority in terms of these statistical metrics. Specifically, the highest MI and FMI indicate our method can effectively integrate source information, \emph{e.g.,}  edges and contrast. Moreover, the significant improvement of VIF implies that our method is consistent with human vision.

\subsection{Comparing with SOTA in Object Detection}

\paragraph{Qualitative comparisons.} 
We visualize the object detection results in Figure~\ref{fig:detection}. At the first row, it can be clearly observed that DIDFuse, TarDAL, and UMFusion lead to the classification or dislocation of the bounding box (\emph{e.g.,} the truck to be detected completely). On the contrary, our method sufficiently preserves the information of categories and results in accurate detection. As for the detection of small targets, \emph{i.e.,}  in the second row, our method detects almost all cars while other methods lead to the omission.

\paragraph{Quantitative comparisons.} 
The quantitative detection results  are presented in Table~\ref{tab:detec}. We can see that our method substantially outperforms other methods in terms of most categories and realizes $10.8\%$ and $9.6\%$ higher detection mAP on the M3FD and Multi-Spectual datasets, respectively.
Apart from detection results, we further conduct the efficiency analysis in Table~\ref{tab:detec}. 
Obviously, benefiting from the dilated convolutions, the minimal time consumption indicates that our method achieves real-time image fusion.

\subsection{Extension to  Multi-modality Segmentation}
Besides comparing with the advanced image fusion, we also provide the  evaluations with three specifically-designed methods: EGFNet~\cite{zhou2021edge}, ABMDRNet~\cite{abmdrnet} and LASNet~\cite{li2022rgb}.

\paragraph{Qualitative comparisons.} 
We select two challenging scenarios to illustrate the remarkable performance compared with eight methods in Figure~\ref{fig:MSPresult}. Two significant advantages can be concluded from these instances. Firstly, our scheme can effectively preserve the complete structure of thermal-sensitive targets (\emph{e.g.,}  pedestrian) with precise prediction.
For example, the pedestrian in the distance cannot be accurately estimated by most methods. Secondly, our scheme also highlights abundant texture details for the normal categories (\emph{e.g.,}  car and curve). Obviously, our method can precisely estimate the structure of cars on both image pairs.
\begin{table}[!htb]
	\footnotesize
	\centering
	\renewcommand\arraystretch{1.1} 
	\setlength{\tabcolsep}{0.7mm}
	\begin{tabular}{c|ccc|ccc}
		\hline
		\multirow{2}{*}{Strategy}  & \multicolumn{3}{c|}{Fusion}                      & \multicolumn{3}{c}{Detection}                   \\ \cline{2-7} 
		& \multicolumn{1}{c|}{\cellcolor{gray!20} {MI$\uparrow$} }& \multicolumn{1}{c|}{\cellcolor{gray!20} {VIF$\uparrow$}} & \cellcolor{gray!20} {FMI$\uparrow$} & \multicolumn{1}{c|}{\cellcolor{gray!20} {Car} }& \multicolumn{1}{c|}{\cellcolor{gray!20} {Person}} & \cellcolor{gray!20} {mAP$\uparrow$} \\ \hline
		\cite{TarDAL}	& \multicolumn{1}{c|}{2.192} & \multicolumn{1}{c|}{0.733} & 0.889 & \multicolumn{1}{c|}{0.736} & \multicolumn{1}{c|}{0.589} &  0.466\\ \hline
		\cite{SeaFusion}	& \multicolumn{1}{c|}{2.210} &\multicolumn{1}{c|}{ 0.751} &\multicolumn{1}{c|}{0.891}    & \multicolumn{1}{c|}{0.660} & \multicolumn{1}{c|}{0.509} &  0.384 \\ \hline
		Proposed	& \multicolumn{1}{c|}{\textbf{2.946}} & \multicolumn{1}{c|}{\textbf{0.913}} & \textbf{0.892} & \multicolumn{1}{c|}{\textbf{0.759}} &\multicolumn{1}{c|}{\textbf{0.598}}& \textbf{{0.531}}   \\ \hline
	\end{tabular}

	\caption{Analyzing the proposed training strategy. }
	\label{tab:strategy1}
\end{table}

\paragraph{Quantitative comparisons.} 
Table~\ref{tab:seg} reports the concrete numerical results on eight categories of the MFNet dataset. Specifically, compared with SeaFusion, which is a segmentation-driven method, our scheme drastically improves 15.2 \% of mIOU.
Noting that, only utilizing one backbone network, our method can achieve comparable performance with specially designed segmentation methods (e.g, LASNet and EGFNet). Especially, we achieve better accuracy in the classification of persons and cars, which is vital for the employment of intelligent vision systems.
\begin{table}[!htb]
	\footnotesize
	\centering
	\renewcommand\arraystretch{1.1} 
	\setlength{\tabcolsep}{1.3mm}
	\begin{tabular}{c|c|c|c|c|c|c}
		\hline
		Task&Metric  &  Manual & EW & GNorm & DWA  & Ours  \\ \hline
		\multirow{3}{*}{Fusion} & \cellcolor{gray!20} {MI$\uparrow$}  & 2.074 &  2.117 &1.696& 1.971& \textbf{2.251}\\ \cline{2-7} 
		&\cellcolor{gray!20} {VIF$\uparrow$} & 0.741 & 0.756 & 0.347 &0.728& \textbf{0.772} \\ \cline{2-7} 
		& \cellcolor{gray!20} {FMI$\uparrow$} & 0.876 &0. 877 & 0.869 &\textbf{0.879}&
		\textbf{0.879} \\ \hline
		\multirow{3}{*}{Detection} & \cellcolor{gray!20} {Car}  & 0.673 &0.670  & 0.683&0.651&\textbf{0.711} \\ \cline{2-7} 
		& \cellcolor{gray!20} {Person} & 0.471 &0.498  & 0.474 &0.509&\textbf{0.536}  \\ \cline{2-7} 
		&  \cellcolor{gray!20} {mAP$\uparrow$} & 0.352 &0.363  & 0.373 &0.362& \textbf{0.428} \\ \hline
	\end{tabular}

	\caption{Analyzing the  aggregation strategy with finite epochs. }
	\label{tab:strategy2}
\end{table}
\begin{table}[htb]
	\footnotesize
	\centering
	\renewcommand\arraystretch{1.1} 
	\setlength{\tabcolsep}{1.3mm}
	\begin{tabular}{c|ccc|ccc}
		\hline
		\multirow{2}{*}{Model} & \multicolumn{3}{c|}{Fusion}                            & \multicolumn{3}{c}{Detection}                            \\ \cline{2-7} 
		& \multicolumn{1}{c|}{\cellcolor{gray!20} {MI$\uparrow$}} & \multicolumn{1}{c|}{\cellcolor{gray!20} {VIF$\uparrow$}} & \cellcolor{gray!20} {FMI$\uparrow$}  & \multicolumn{1}{c|}{\cellcolor{gray!20} {Car}} & \multicolumn{1}{c|}{\cellcolor{gray!20} {Person}} & \cellcolor{gray!20} {mAP$\uparrow$}  \\ \hline
		TarDAL$_\mathtt{J}$	& \multicolumn{1}{c|}{\textbf{2.558}} & \multicolumn{1}{c|}{\textbf{0.661}} & 0.825 & \multicolumn{1}{c|}{{0.652}} & \multicolumn{1}{c|}{{0.515}} & 0.404 \\ \hline
		TarDAL$_\mathtt{P}$	& \multicolumn{1}{c|}{2.465} & \multicolumn{1}{c|}{0.614} & \textbf{0.843}  & \multicolumn{1}{c|}{\textbf{0.714}} & \multicolumn{1}{c|}{\textbf{0.555}} &\textbf{0.440}  \\ \hline
		UMFusion$_\mathtt{J}$	& \multicolumn{1}{c|}{2.470} & \multicolumn{1}{c|}{0.585} & 0.855 & \multicolumn{1}{c|}{0.692} & \multicolumn{1}{c|}{0.518} & 0.408 \\ \hline
		UMFusion$_\mathtt{P}$	& \multicolumn{1}{c|}{\textbf{2.999}} & \multicolumn{1}{c|}{\textbf{0.653}} & \textbf{0.881} & \multicolumn{1}{c|}{\textbf{0.712}} & \multicolumn{1}{c|}{\textbf{0.523}} & \textbf{0.452} \\ \hline
	\end{tabular}
	
	\caption{Evaluating of generalization ability on M3FD benchmark. }
	\label{tab:strategy4}
\end{table}
\subsection{Ablation Studies}

\begin{figure}[htb]
	\footnotesize
	\centering
	\setlength{\tabcolsep}{1pt}
	\begin{tabular}{cccc}
		\includegraphics[width=0.105\textwidth,height=0.07\textheight]{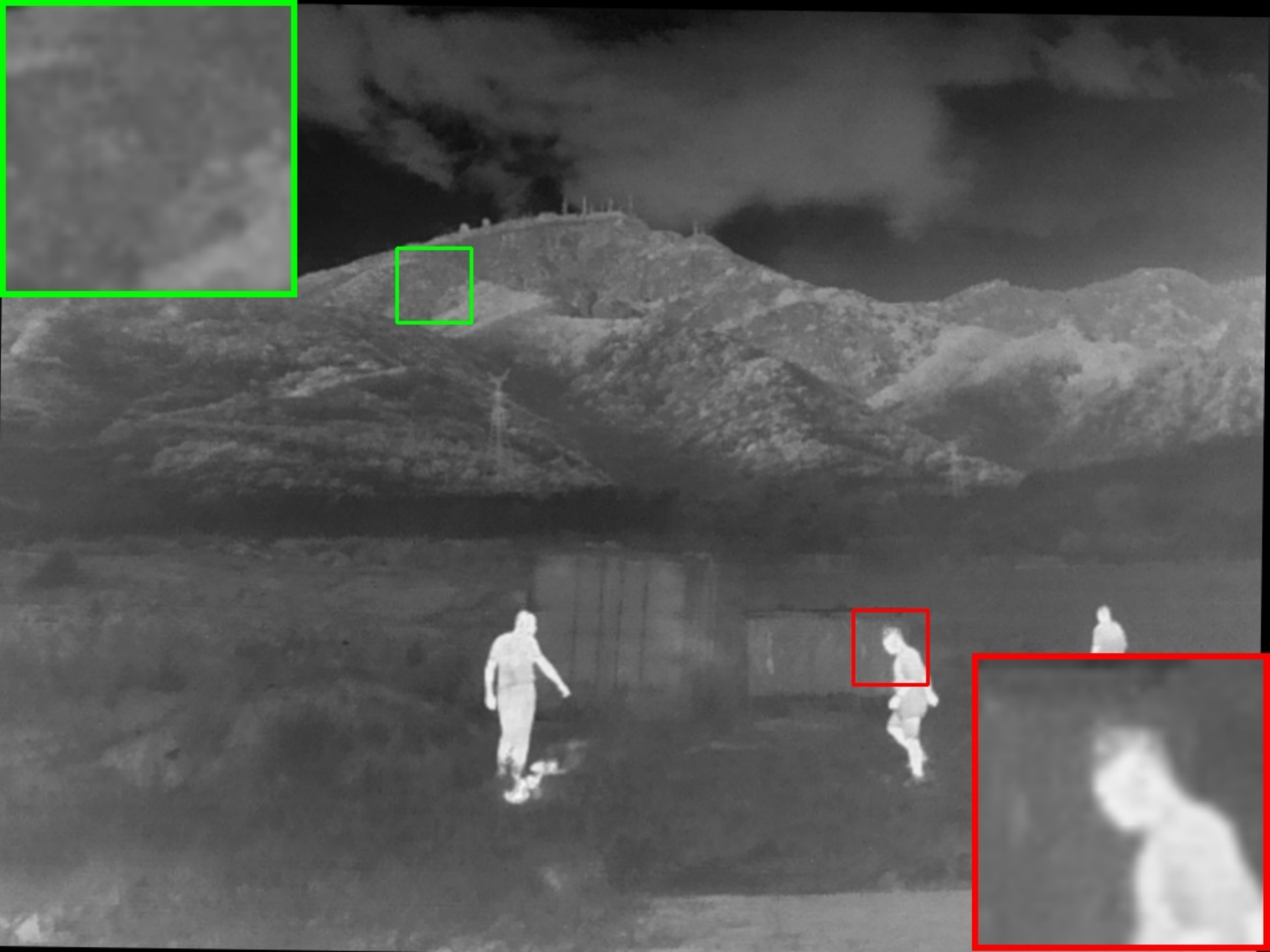}
		&\includegraphics[width=0.105\textwidth,height=0.07\textheight]{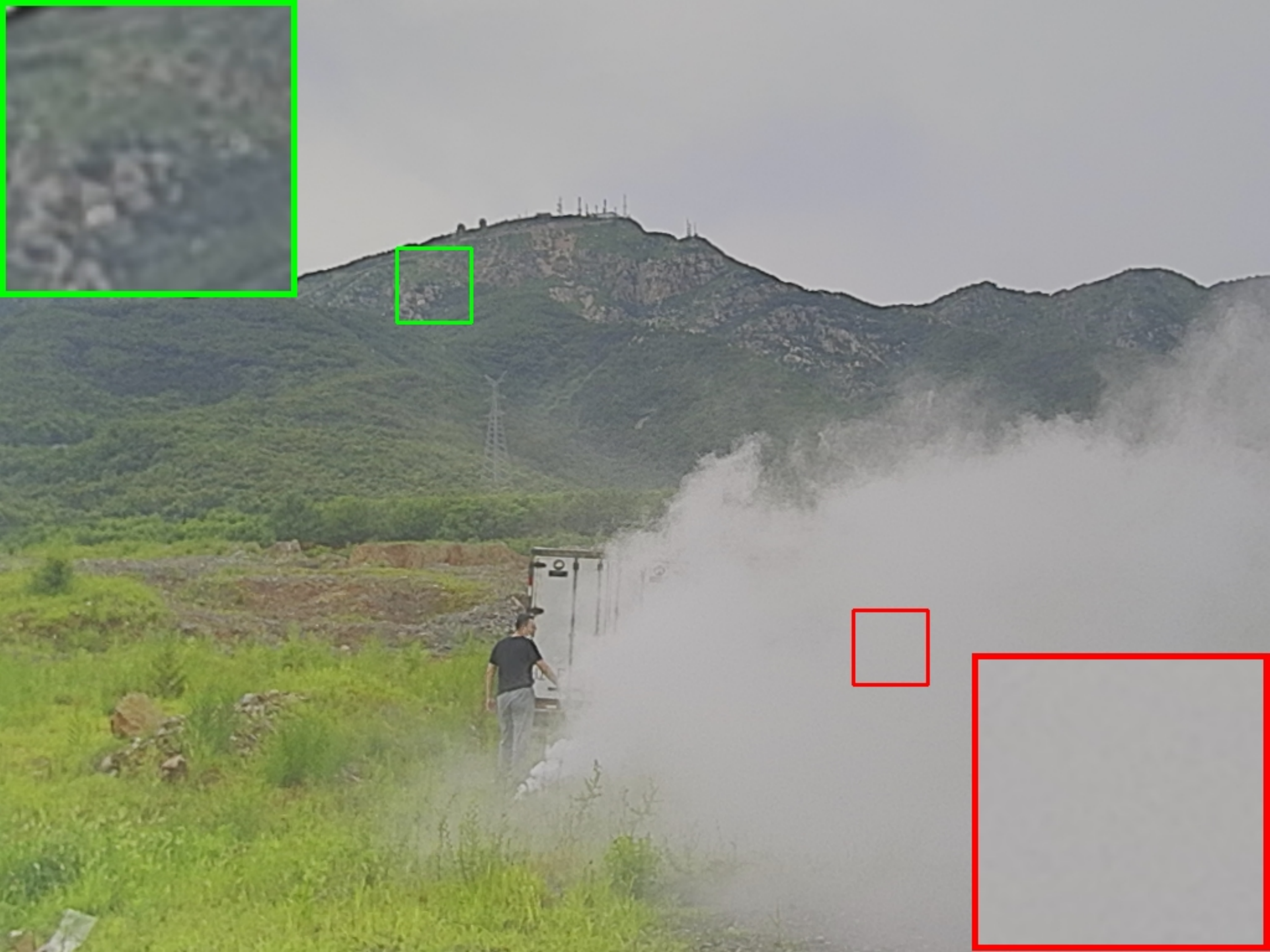}
		&\includegraphics[width=0.105\textwidth,height=0.07\textheight]{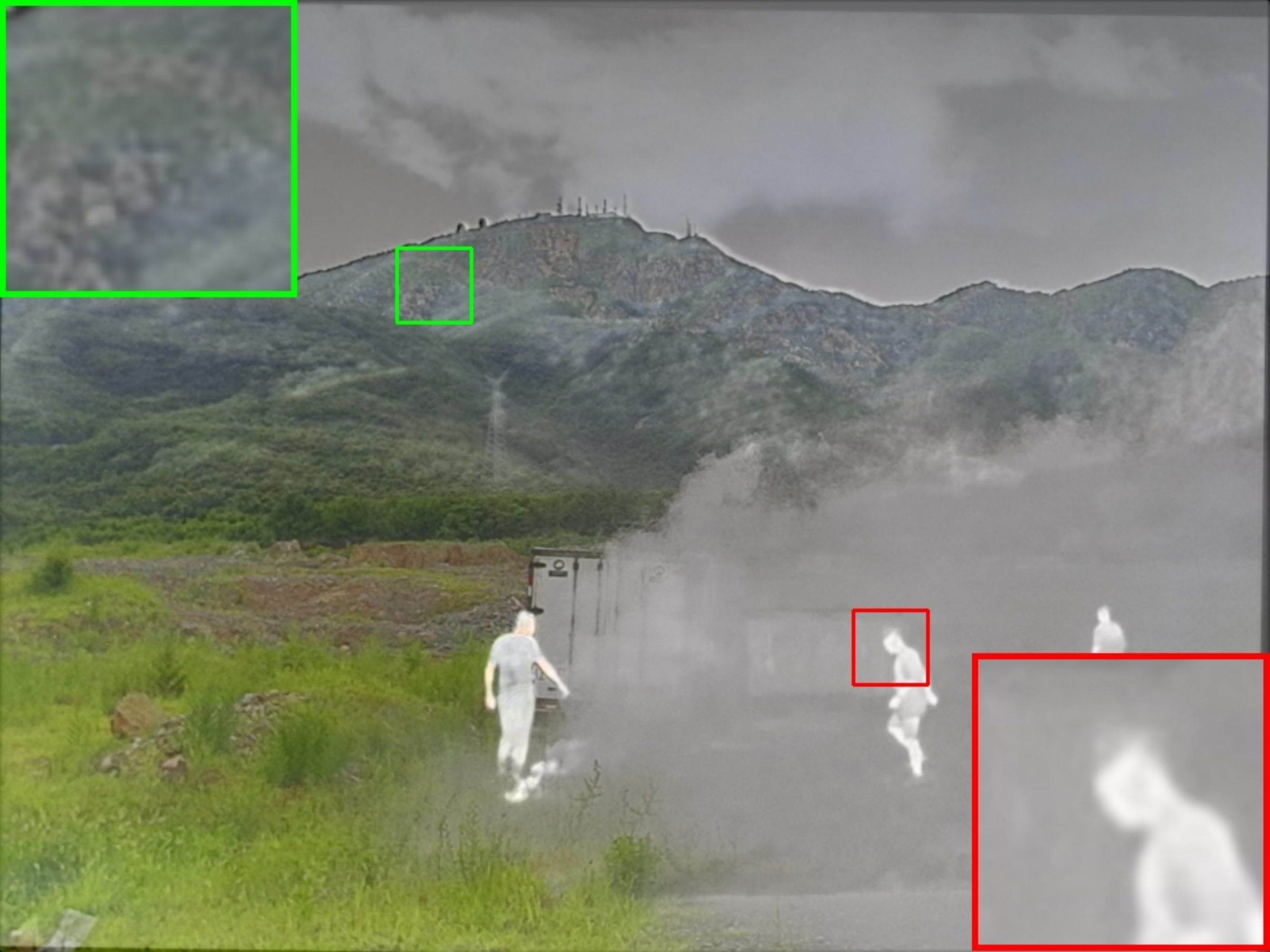}
		&\includegraphics[width=0.105\textwidth,height=0.07\textheight]{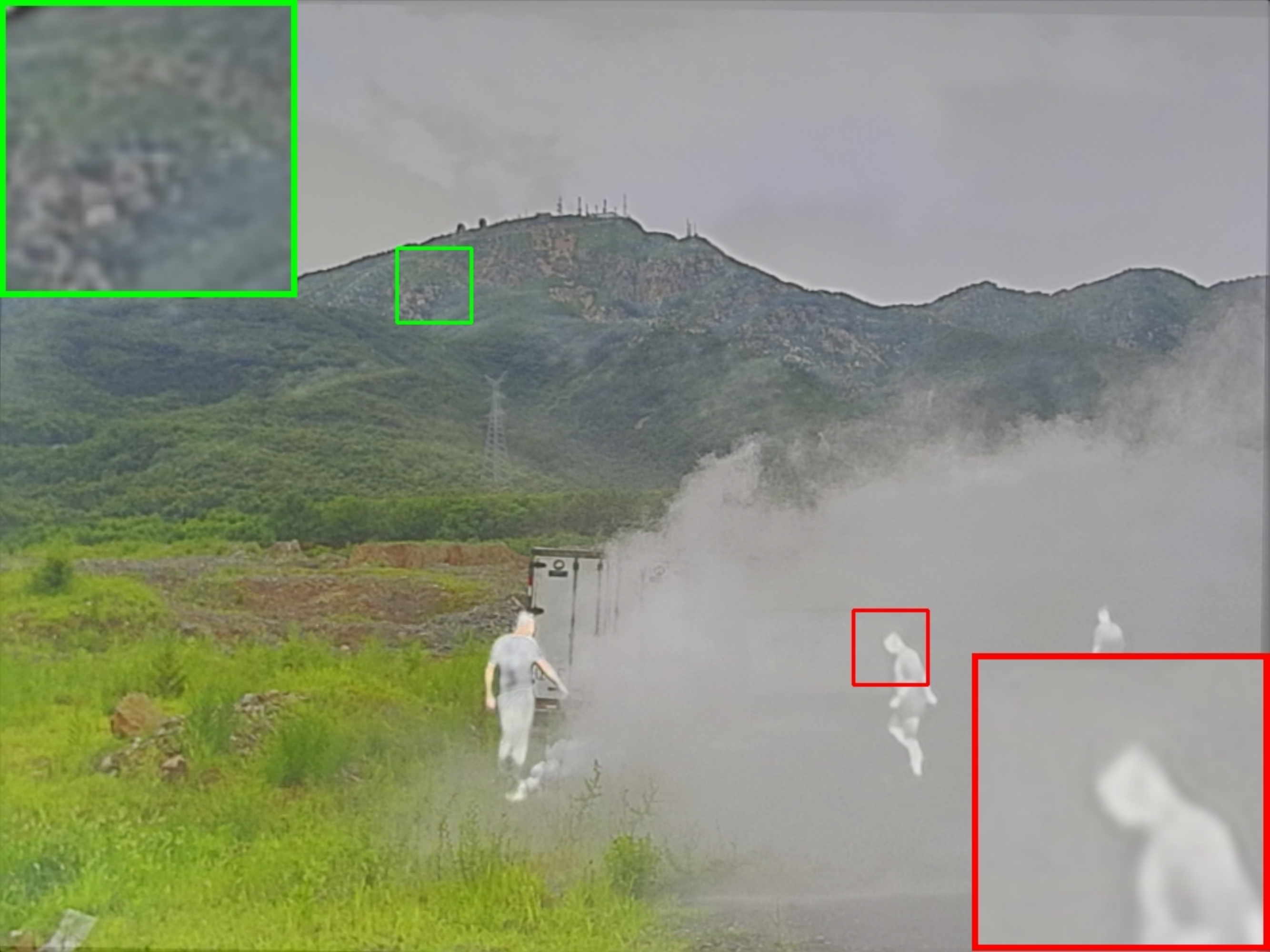}\\
		\includegraphics[width=0.105\textwidth,height=0.07\textheight]{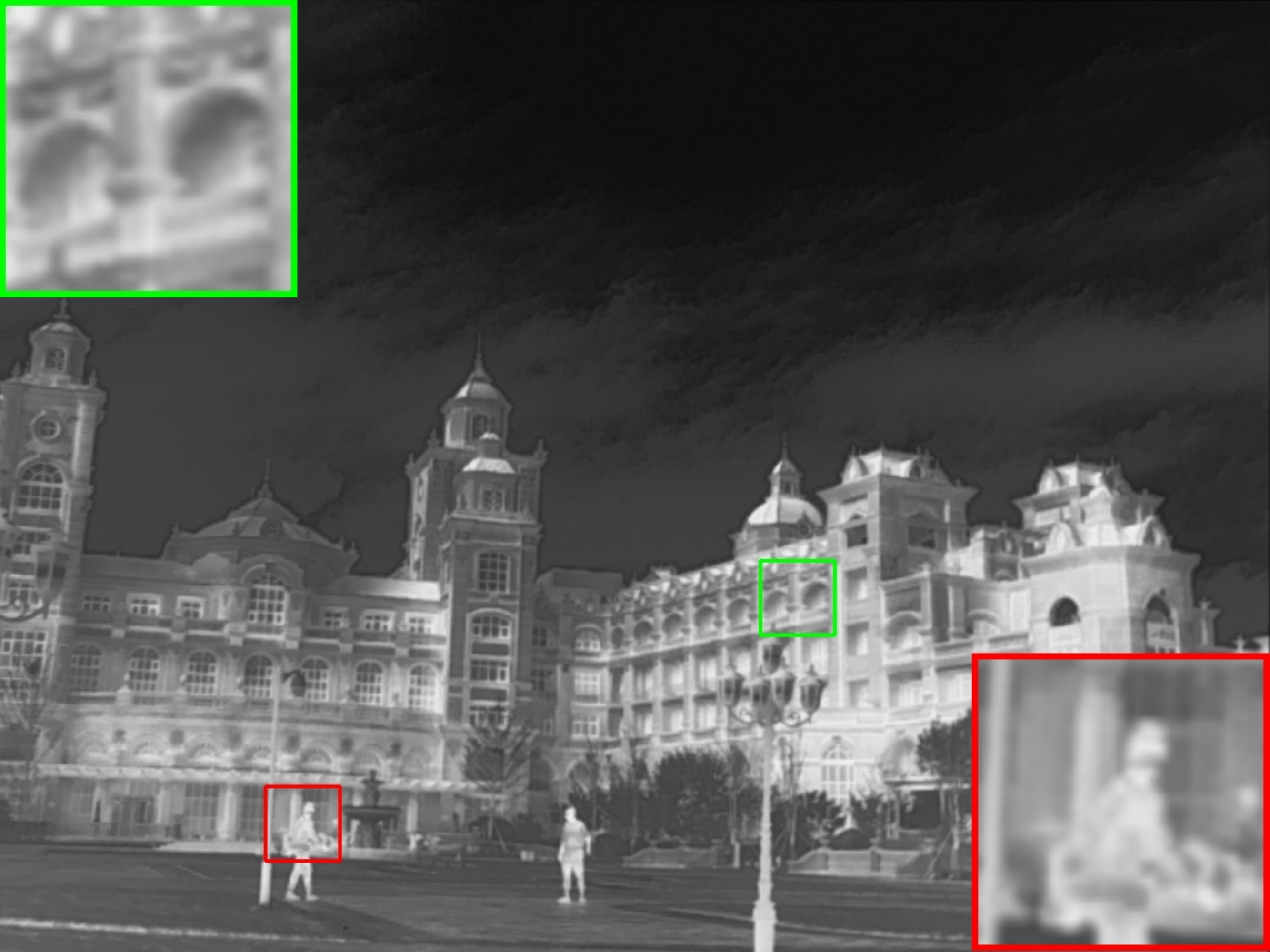}
		&\includegraphics[width=0.105\textwidth,height=0.07\textheight]{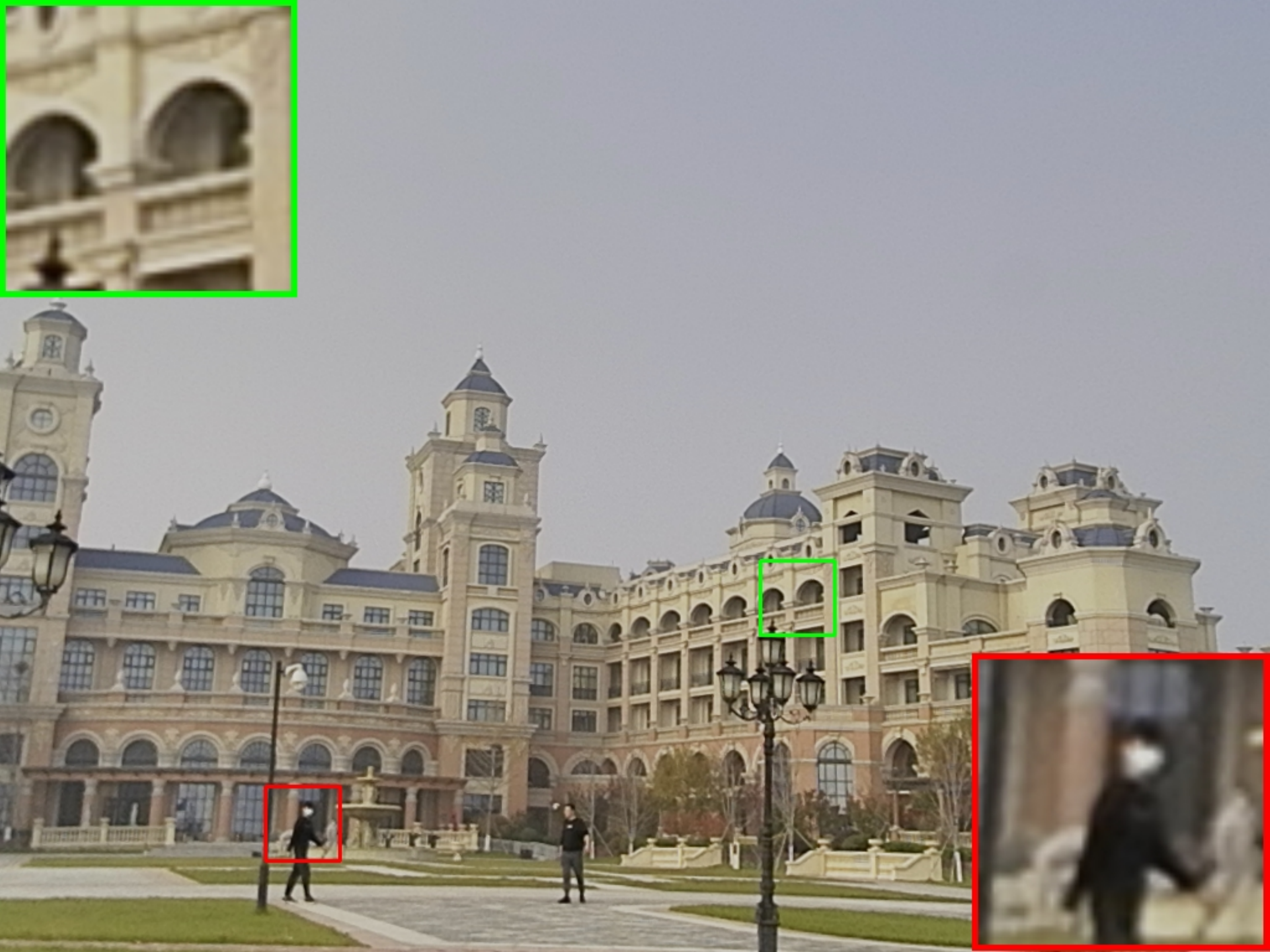}
		&\includegraphics[width=0.105\textwidth,height=0.07\textheight]{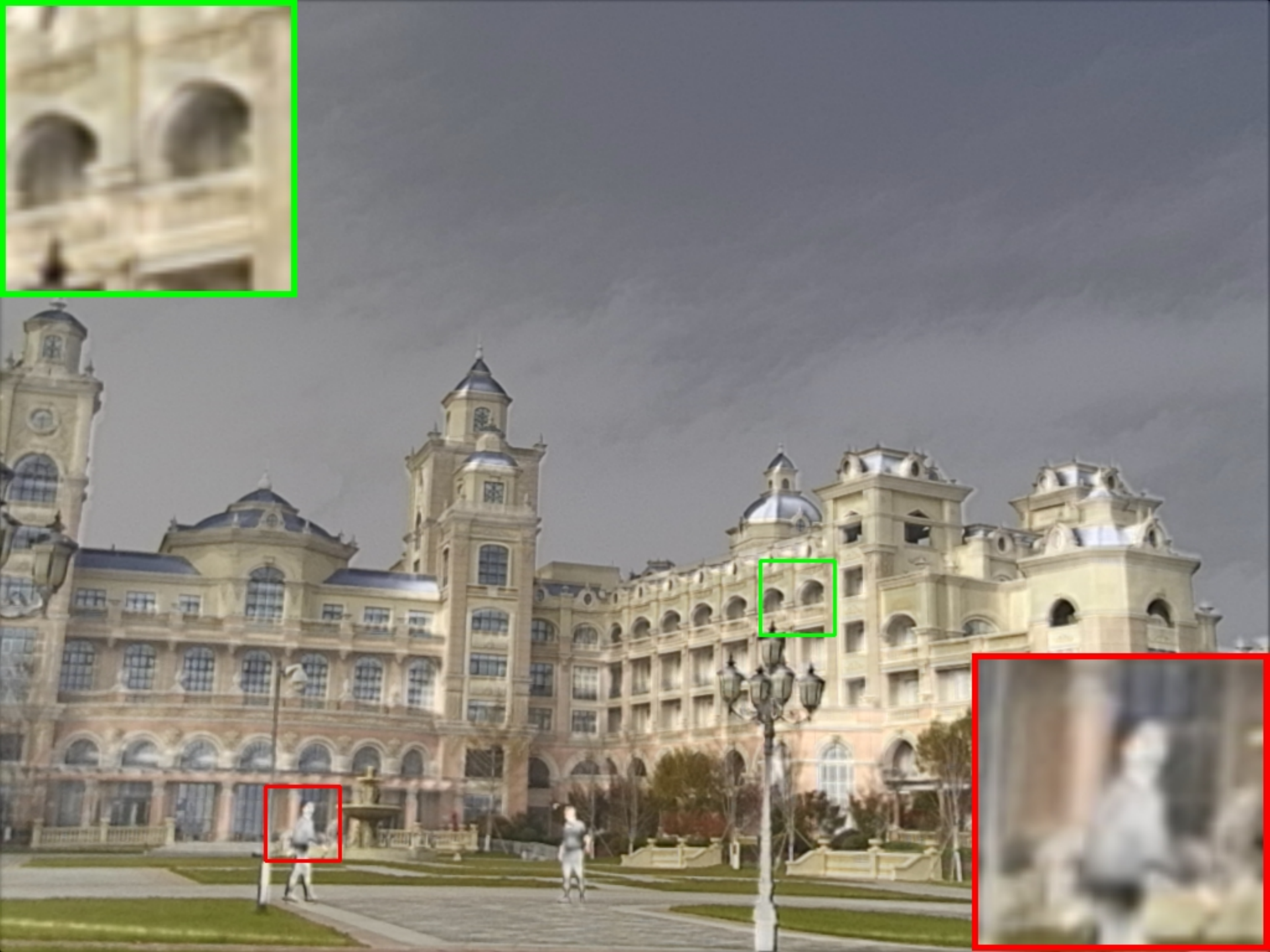}
		&\includegraphics[width=0.105\textwidth,height=0.07\textheight]{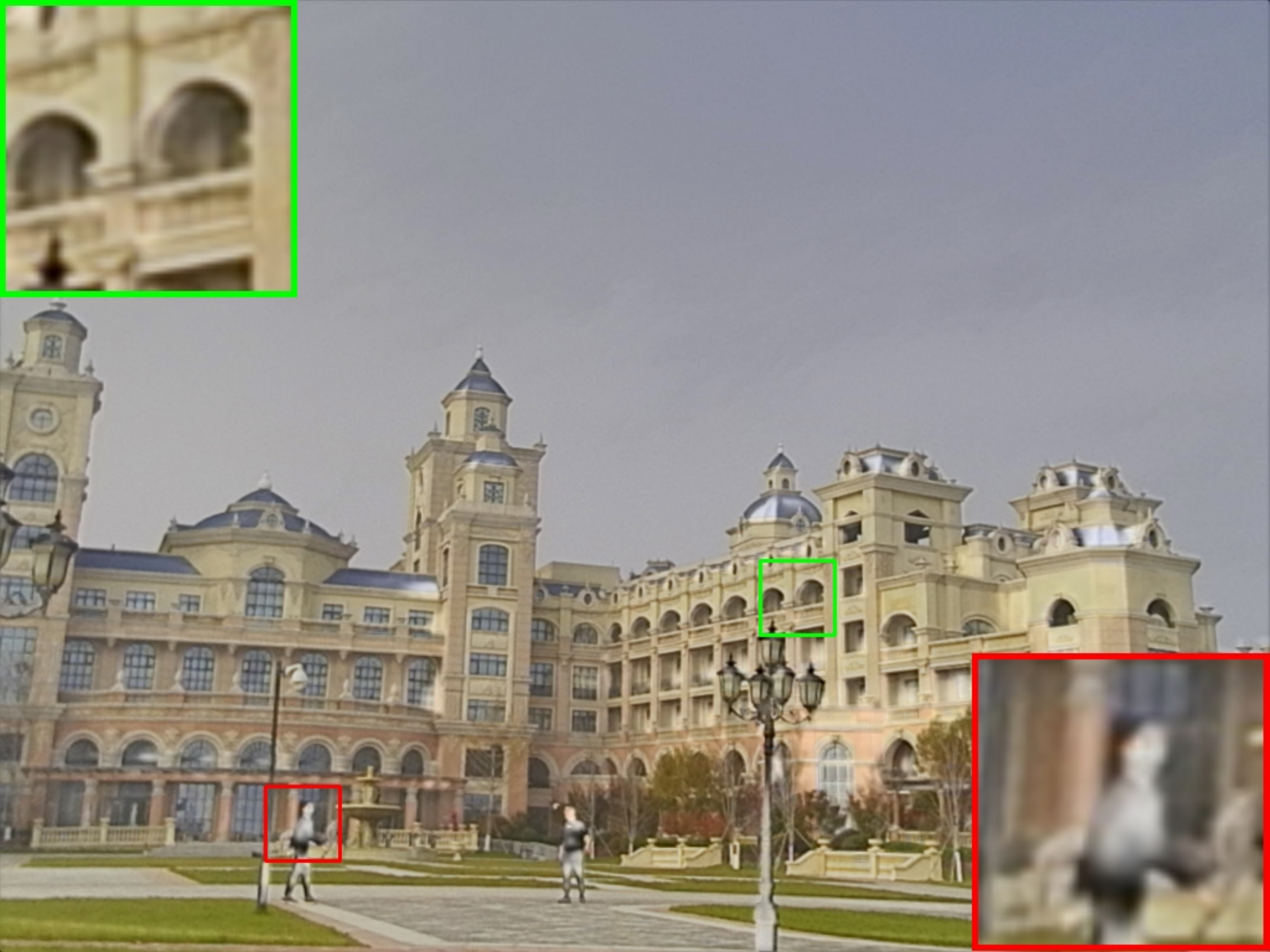}
		\\	
		Infrared & Visible & DgF & SgF 
	\end{tabular}

	\caption{Visual differences of  fusion guided by perception tasks. }
	\label{fig:lamda}
\end{figure}
\paragraph{Effects of proposed training strategy.} To demonstrate the effectiveness of the bi-level learning paradigm, we conduct the comparison with two mainstream training strategies (\emph{i.e.,}  unrolled training~\cite{TarDAL} and adaptive loop training~\cite{SeaFusion}). 
Table\ref{tab:strategy1} reports the 
numerical results of image fusion and detection based on these strategies. Both existing strategies are concrete on one task and neglect another. We can observe that the proposed strategy achieves consistent performance for the two scene parsing tasks.

\paragraph{Evaluating of dynamic aggregation strategy.} In order to evaluate the effectiveness of  dynamic weights, we provide other four classical adjustment ways, \emph{i.e.,}  manual-design, equal weighting, gradient normalization~\cite{chen2018gradnorm} and DWA~\cite{liu2019end}. The numerical results (trained with 10 epochs) are reported in Table~\ref{tab:strategy2}. The random aggregation can effectively avoid the local optimum with higher performance compared with representative multi-task learning and manual adjustment.

\begin{table}[h]
	\footnotesize
	\centering
	\renewcommand\arraystretch{1.1} 
	\setlength{\tabcolsep}{1.4mm}
	\begin{tabular}{c|ccc|ccc}
		\hline
		\multirow{2}{*}{Guided Task}  & \multicolumn{3}{c|}{TNO}                      & \multicolumn{3}{c}{M3FD}                   \\ \cline{2-7} 
		&  \multicolumn{1}{c|}{\cellcolor{gray!20} {MI$\uparrow$}} & \multicolumn{1}{c|}{\cellcolor{gray!20} {VIF$\uparrow$}} & \cellcolor{gray!20} {FMI$\uparrow$} &  \multicolumn{1}{|c|}{\cellcolor{gray!20} {MI$\uparrow$}} & \multicolumn{1}{c|}{\cellcolor{gray!20} {VIF$\uparrow$}} & \cellcolor{gray!20} {FMI$\uparrow$} \\ \hline
		Detection	& \multicolumn{1}{c|}{\textbf{2.946}} & \multicolumn{1}{c|}{\textbf{0.913}} & \textbf{0.892} & \multicolumn{1}{c|}{3.033} & \multicolumn{1}{c|}{0.699} &0.863  \\ \hline
		Segmentation	& \multicolumn{1}{c|}{2.902} & \multicolumn{1}{c|}{0.745} & 0.889 & \multicolumn{1}{c|}{\textbf{3.482}} & \multicolumn{1}{c|}{\textbf{0.781}} &  \textbf{0.864}\\ \hline
	\end{tabular}

	\caption{Visual influences of perception tasks. }
	\label{tab:strategy3}
\end{table}
\paragraph{Evaluating of generalization ability.} The proposed bi-level dynamic learning is a generalized scheme, which is network-agnostic. Table~\ref{tab:strategy4} illustrates the significant improvement for  advanced
fusion networks (\emph{i.e.,}  TarDAL and UMFusion) using our training strategy. Subscripts ``$\mathtt{J}$'' and ``$\mathtt{P}$'' denote  the joint and proposed training strategies.

\paragraph{Influences of perception tasks for image fusion.} We also analyze the 
influence of diverse perception guidance for image fusion in Table~\ref{tab:strategy3} and Figure~\ref{fig:lamda} respectively. We can observe that both quantitative results are higher than current advanced methods. Detection-guided fusion (denoted as ``DgF'') realizes better results on TNO, which contain abundant targets with high contrast. Segmentation-guided fusion (denoted as ``SgF'') achieves a significant improvement on M3FD, which is with vivid texture details. From Figure~\ref{fig:lamda}, we can observe DgF trends to preserve the salient targets (\emph{e.g.,}  person). The results of SgF prefer to maintain the scene background with obvious structural information for  pixel-wise classification.

\section{Conclusion}
In this paper, a hierarchical dual-task deep model was proposed to bridge multi-modality image fusion and semantic perception jointly. We presented a bi-level formulation to depict the coupled relationship, to realize mutual promotion. Then a dynamic aggregation strategy was proposed to combine the gradients from distinct tasks to achieve the comprehensive image fusion. Extensive experiments demonstrated our method is superior to address diverse vision tasks.

\appendix
%
%

\section*{Acknowledgments}

This work is partially supported by the National Key R\&D
Program of China (2020YFB1313503), the National Natural
Science Foundation of China (U22B2052),
the Fundamental Research Funds for the Central Universities
and the Major Key Project of PCL (PCL2021A12).

\bibliographystyle{named}
\bibliography{egbib}

\end{document}